# How Well Can AI Build SD Models?

By: Billy Schoenberg, Davidson Girard, Saras Chung, Ellen O'Neill, Janet Velasquez, Sara Metcalf

bschoenberg@iseesystems.com
davidson@skipdesigned.com
saras@skipdesigned.com
ellen@skipdesigned.com
janet@skipdesigned.com
smetcalf@buffalo.edu

**Short Summary**

> **Introduction:**
> As system dynamics (SD) embraces automation, AI offers efficiency but risks bias from missing data and flawed models. Models that omit multiple perspectives and data threaten model quality, whether created by humans or with the assistance of AI. To reduce uncertainty about how well AI can build SD models, we introduce two metrics for evaluation of AI-generated causal maps: technical correctness (causal translation) and adherence to instructions (conformance).
>
> **Approach:**
> We developed an open source project called sd-ai to provide a basis for collaboration in the SD community, aiming to fully harness the potential of AI based tools like ChatGPT for dynamic modeling. Additionally, we created an evaluation theory along with a comprehensive suite of tests designed to evaluate any such tools developed within the sd-ai ecosystem.
>
> **Results:**
> We tested 11 different LLMs on their ability to do causal translation as well as conform to user instruction. gpt-4.5-preview was the top performer, scoring 92.9% overall, excelling in both tasks. o1 scored 100% in causal translation. gpt-4o identified all causal links but struggled with positive polarity in decreasing terms. While gpt-4.5-preview and o1 are most accurate, gpt-4o is the cheapest.
>
> **Discussion:**
> Causal translation and conformance tests applied to the sd-ai engine reveal significant variations across lLLMs, underscoring the need for continued evaluation to ensure responsible development of AI tools for dynamic modeling. To address this, an open collaboration among tool developers, modelers, and stakeholders is launched to standardize measures for evaluating the capacity of AI tools to improve the modeling process.

# 1. Introduction

With the advent of automation, the phrase, "everybody thinking in systems" is even more possible now than ever before. Expert modelers may no longer be the hard barrier to entry for building qualitative system dynamics (SD) models as they were only a few short years ago (Black & Greer, 2024). The rapid development of generative artificial intelligence (AI) has come to the system dynamics field (du Plooy & Oosthuizen, 2023, Giabbanelli et al., 2023, Armenia et al., 2024, Hosseinichimeh et al., 2024; Ghaffarzadegan et al., 2024; Jalali & Akhavan, 2024; Liu & Keith, 2024; Veldhuis et al., 2024; Giabbanelli et al., 2025; Hu, 2025).  Developments in software, via advancements in data science and machine learning methods, have produced tools like ChatGPT, which help people who are untrained in the formal methods of system dynamics to generate qualitative models as causal maps or causal loop diagrams (CLDs) (Hosseinichimeh et al., 2024). Even if you ethically object against it, automation using AI is here and its use appears to be only growing (Xu et al., 2024). While AI has the potential to expand the field of system dynamics by attracting more practitioners, fewer people may actually be engaged in the *process* of systems thinking as a result. As automated tools make the process of modeling more accessible, it is imperative for the field to ensure the development of high quality system dynamics models using AI.

In this paper, we propose that care should be taken when applying AI, namely Large Language Models (LLMs), to system dynamics modeling. Results should be trustworthy, useful, high quality, and adhere to the well established methodological practices of the field. First, we outline the benefits and risks of using AI in system dynamics and make the case for an open-source platform to conduct this work transparently for the purpose of evaluation. We then test the technical accuracy of a new tool, "sd-ai," and report the performance of our work when we use different LLMs to generate results. Finally, we close with a framework and institutional organization for the continued evaluation and quality improvement of our efforts to usher system dynamics into the age of AI.

## 1.1 The adoption of AI to improve system dynamics modeling

Whether you believe in the power of AI to enhance human thriving or are concerned about its implications to society, AI is here and integrated into many parts of the everyday human experience (Hirsch-Kreinsen, 2024). From the predictive text suggested by your smartphone to the movies that are recommended through your streaming provider, almost every aspect of modern human lives are influenced by predictive intelligence. This pervasive presence of AI doesn't just offer convenience, it also shapes how we perceive and interact with the world. As algorithms curate the content we consume, they create what some scholars call an "information cocoon," filtering our experiences and subtly reinforcing certain perspectives while obscuring others (Xu, Wang, & Zhang, 2024). In this way, AI is not just a passive tool but an active force in constructing our understanding of reality.

The growing reliance on AI-driven predictions has significant implications for system dynamics where predictive analytics and causal inference is central to our work. As AI reshapes how we process and interpret information, it also presents opportunities for integrating automation into our modeling methods. System dynamics was once described as a field that was "theory-rich, but data poor" (Pruyt et al., 2014), but now we find ourselves in a paradoxical scenario where the rise of big data and LLMs make vast amounts of information more accessible with no guarantee of theoretical rigor. Sterman (2018) quoted Jay Forrester's view that the advent of "big data" and machine learning will likely create unprecedented opportunities for dynamic modelers. Forrester's viewpoint suggested that while ML and big data could enhance modeling capabilities by providing richer empirical inputs, they should not replace the fundamental principles of system dynamics – including modelers. These fields/technologies should, instead, be leveraged to improve calibration, validation, and refinement of dynamic models. The challenge, according to Forrester via Sterman (2018), is to ensure that an over-reliance on data does not lead to models that lack causal depth or structural insight. In short, system dynamics practitioners can (must?) thoughtfully integrate big data – and by extension, AI – into their work using tools that complement traditional methodologies.

As a field, system dynamics is starting to embrace the idea of building models using AI. The areas of interest for AI in system dynamics have been vast and varied. From exploring applications (e.g., Armenia et al., 2024) to experimenting with various aspects of automation (e.g., Hu, 2024), system dynamicists are engaging, building, and creating new ideas with AI (e.g., Jalali & Akhavan, 2024; du Plooy & Oosthuizen, 2023). As we continue to play with the development of AI tools for easier and more efficient modeling, there are a number of opportunities and risks that are becoming apparent.

## 1.2 The opportunities of using automation for system dynamics modeling

In the field of system dynamics, the push to integrate AI into our model building methods (separate from validation or calibration) can be distilled to a few key motivators: 1) to expedite the typically slow research and mapping process, 2) to increase the robustness or inclusion of large sets of data through the use of LLMs, 3) to enhance impact by staying abreast of developments in computational power. For instance, the use of AI has been noted to support the generation of dynamic causal theories in order to significantly expedite research processes (Ghaffarzadegan et al., 2024; Liu & Keith, 2024; Veldhuis et al., 2024). By integrating AI into the modeling process via LLMs, we gain efficiencies with code generation, story creation and certain reasoning tasks, demonstrating promise in translating dynamic hypotheses into CLDs without explicitly drawing loops and relationships one by one (Liu & Keith, 2024). There may be utility for AI to help novice modelers exponentially reduce the barriers they face engaging with the more sophisticated methods of simulation modeling and experimentation. On the flipside, reduced barriers to creating system dynamics models with AI may lead to lower quality models and learning.

One of the greatest threats to adopting system dynamics more widely in practice is the creation and dissemination of low-quality models (Forrester, 2007; Homer, 2013). A frequently asked question is how we know if the model is right. A series of methods to build confidence in models has been developed, including Barlas' seminal work on model validation (Barlas, 1996), Senge & Forrester's work on model testing (Senge & Forrester, 1980), Sterman's guide for confidence building (Sterman, 2000), and Rahmandad & Sterman's guidelines on model documentation and testing (Rahmandad & Sterman, 2012). While these resources are available to the human modeler, the same confidence testing methodology is not necessarily inherent in automated system dynamics models. Nor is it done iteratively, as best practice may suggest (e.g., Sterman, 2000; Schwaninger, 2019). How do we know what went into creating a model and whether the structure fits various confidence building tests? This is a key question we should be addressing in the next phase of automated system dynamics.

Despite the warnings against the use of AI, there is a sense that the field may be left behind if it fails to adopt these technological innovations. Jay Forrester himself was a contributor to the invention of the digital computer, perhaps a fear-inducing development in an era that was devoid of the Internet and many of the other technological advancements we now have which were only possible because of the digital computer. During a time when computational power was limited, the field of system dynamics emerged to support decision-making in complex systems by making mathematically explicit the causal feedback mechanisms and behavior over time (Forrester, 1961). Naturally, one would think the field would then embrace the development of AI to further our methods.

Moreover, we should be asking, "To what end?" How can we as a field leverage improvements in data science and computational power to ultimately move from theorizing to doing? Forrester encouraged the field not simply to advance our methods for the sake of technological innovation, but for real-world impact. To do so, however, requires an understanding of our limitations and the ways in which new technologies may make the impact we fear by changing the world in the opposite direction we hope. This intuitive "leverage point" around the use of AI to build system dynamics models could be argued to make things worse by "pushing it in the wrong direction" as Donella Meadows warned in her seminal paper on leverage points (Meadows, 1999). In her depiction of typical ways in which decision-makers make change, Meadows (1999) describes subsidies and taxes (and other forms of feedback loops) that are meant to alleviate negative outcomes for the world but instead do just the opposite. The same could be true for our field's growing emphasis on AI. We are perhaps destined to fall in the same trap if we compete in the AI arms race by developing tools without realizing the harm they may do inherently by creating them.

## 1.3 The risks of using automation for system dynamics modeling

A taxonomy of risks was recently published that outlines the various ethical and social risks that automated models may pose to the world. Weidinger et al. (2023) identified twenty-one such risks situated in a taxonomy of six different risk areas: 1) discrimination, hate speech and exclusion, 2) information hazards, 3) misinformation harms, 4) malicious uses, 5) human-computer interaction harms, and 6) environmental and socioeconomic harms. These areas have serious implications for the use of AI in society, including the adaptation of these tools for system dynamics models.

For instance, bias has been often depicted as an obvious outcome of AI. The most widely accepted issue of bias with AI in LLMs for social challenges is the systematic omission of people groups and their perspectives from information on which LLMs (and other AI tools) are trained. Forrester argued that both numerical data and "soft" or unquantified variables should be included in models if they are important to the purpose, that quantifiable data are only a tiny fraction of the relevant data needed to develop a model and emphasized the importance of written material in the "mental database", which consists of mental models, beliefs, perceptions, and attitudes of actors in a system (Sterman, 2018). Sterman (2018) goes further to say that people most often misunderstood Forrester's emphasis on the mental databases of people to mean just to go and "talk to people," instead pointing to an ethnographic or cultural anthropological approach to engaging in a system. "...first hand knowledge can be obtained only by living and working where decisions are made and by watching and talking with those who run the economic [or any] system" (Forrester, 1980, p. 557). AI poses a debatable challenge for this type of engagement as we rely on automation of existing data to do the work that a modeler may do to engage in and deeply understand and depict a system.

Unfortunately, there are numerous groups of people whose data and observations about causality may be systematically missing due to their lack of sharing and building knowledge on the web: children, older adults, indigenous societies, and people in regions or countries where the Internet has not been pervasively deployed. Martin et al. (2020) notes that when models exclude causal inferences of key stakeholders, especially people who have lived experience in a complex system, there is a systematic bias that may limit the usefulness of a model and argued the use of participatory model building practices as a way to beat the bias that results from machine learning.

Relatedly, in her book, *Race after Technology*, Ruha Benjamin (2019) argues that though perhaps even unintentionally, automation has the potential to cover up, accelerate and deepen discrimination while appearing neutral and even benevolent compared to racism in previous eras. She presents the concept of the "New Jim Code" and how discriminatory designs encode inequity and amplify racial hierarchies by repeating social divisions or aiming to fix racial bias but doing the opposite. Bias may also result from the quality of data which is largely made without review from the public, which are perhaps not always legal or accurate sources. "Hallucinations" or fictional data created by AI are a widely known challenge that may result in faulty knowledge and fictional data.

The issue of misinformation or disinformation on the web is a rampant issue that is difficult to regulate. With the added integration of AI, the potential for misinformation to make it into our system dynamics models is especially troubling (Shin, 2024; Monteigh, Glenn et al., 2024). The web is crawling with conflicting perspectives and all the expected and surprising human biases on the internet. What someone may publish as truth, another person may question and denounce. These oppositional sources of information are not necessarily sifted out of automation, leading to potential erroneous models that repeat biases, create prejudices and explicit efforts to deceive or mislead others.

Additionally, models are increasingly trained on the output from other models or from privately held data sources. Many companies at the frontiers of data science, machine learning, and LLM development seem to take pride in their secret data recipe. Knowing the data doesn't tell you how models will interpret the data into their responses, because underlying architecture is optimizing for correct token prediction, not for directly storing any specific knowledge or facts. This can be problematic from a system dynamics point-of-view, where understanding how a model works is part of transformation. Black and Greer's (2024) work on boundary objects should remind the field that system dynamics models are only as useful as how they can mind the abstraction gap from models to action. The less we know about the way our models are construed, the less useful they may be on the ground to make a real world impact.

Moving forward, we need ways to integrate human cognition and revision with automated models. How do we embrace automation with ways for people to integrate new voices and avoid echo chambers/endogenous modeling data? Another challenge posed by the automation of system dynamics models at this stage is the potential to rely on qualitative modeling without moving into quantitative simulation, a decades-old challenge that may be exacerbated by easier modeling approaches (e.g., Richardson, 1999). These are important challenges. However, even before we are able to tackle this issue, we must work on the fundamentals: creating technically accurate system dynamics models with automation using LLMs. Post-training regimes add an additional layer of functionality and safety to models that can impact their accurate adherence to underlying datasets.

## 1.4 The technical accuracy of automated system dynamics models

Due to the many opportunities and risks involved in the automation of system dynamics models and the need for high quality models that are useful, it is important to determine which LLMs can accurately transcribe causal relationships from written English to a system dynamics based relationship form. Useful models can be made with efficiency when the data that are used to train current AI systems are holistic and the diagrammatic notations appropriately depict polarities, causal direction, and loop structures.

The act of drawing qualitative models using causal maps or CLDs has been difficult for novices to perform successfully, sometimes taking many repetitions of training and mentorship. Richardson (1999), noting this as a challenge, suggested the development of a compendium of principles and rules for creating causal maps from texts, but noted the creation of coding rules would be difficult and controversial, perhaps because there are various perspectives on the issue. For instance, the polarity notation of a CLD can be confusing when a plus means the origin variable decreases, the resulting variable also decreases. Many students of systems thinking who are forraying into causal loop diagramming notation are often confused by the counterintuitive labeling of the (+) polarity in that situation. The same goes for the (-) notation where an increase in one variable leads to a decrease in the resulting variable. From completely excluding these notations to offering alternative ways to depict causality through same (S) or opposite (O) notation, there have been a number of efforts to make the causal mapping of a system dynamics model easier or more accessible (Barbrook-Johnson & Penn, 2022), the language humans use to describe these relationships can be confusing. Additionally, causality is complicated to determine if using data that is retrieved from the web. Even recently published articles and websites may confuse or insinuate causation based on correlational relationships, creating more noise that needs to be sorted through in the process of building a system dynamics model.

In the new and rapidly developing technique of using LLMs to develop CLDs, few evaluation techniques exist to support our understanding of whether any LLM is better than another in generating accurate CLDs. We propose the following as the first steps towards building a field that integrates high quality system dynamics modeling with AI:

1) Set best practices for the field for including diverse knowledge "in-context" for LLMs and working to make sure our software tooling supports these best practices.
2) Establish an industry standard test set for evaluating accuracy of LLM generated models.
3) Educate current and next generation modelers on appropriate and inappropriate uses of AI.

The next section introduces the groundwork that has been done by the field and pathways to achieve the goals outlined above.

## 2. The sd-ai project

To achieve the goals and address the risks of growing the field of system dynamics with AI, transparent dialogue and development will be needed in our community. Based on conversations with people in the field, it became clear that creating an open source platform to encourage collaboration among researchers both within and outside of the system dynamics community could help us further our field's engagement with AI. We hope that this sd-ai platform facilitates the technical development of an open source web-based service to leverage the power of generative AI and expand the use of systems thinking in society.

Open source was key to achieve our goals of increasing cooperation among researchers. By releasing sd-ai under a permissive open source license (the MIT license), the entire researcher and user community will be able to benefit from everyone's work on the project. These benefits include being able to freely use the tools as well as transparency into others' underlying approaches, whether that be prompts on top of off-the-shelf LLMs or their system design, and technical implementation details. Likewise, by developing a web service which interacts via a REST (Representational State Transfer) based API, we maximize the reach of the sd-ai platform for three key reasons: first, because the existing tools within the system dynamics field (e.g. Stella, CoModel) are able to interact in a seamless manner with web services; second, because the software engineering challenges of working with REST-based APIs are well understood (Neumann et al., 2018); and third, because REST-based APIs are easily consumable from a wide variety of program languages (C++, Python, R etc) operating on a wide variety of systems (e.g., desktop, cloud, mobile devices). Another key architectural decision was the choice to develop the sd-ai application using Node.js due to its ease of hosting, ease of development, and general popularity within web-development software engineering circles (Chaniotis et al., 2014). By working with standard well-known tools, we hope to lower barriers to entry for future software developers interested in working on the sd-ai project.

As seen in Figure 1, the sd-ai project is set up as a generic platform to support the development of what we've termed "engines," which do the work of communicating with externally hosted LLMs to generate model content and are the area where we expect researchers within the community to make their primary contributions. The flow of information from the user is marked in black, the flow of information to the user is marked in blue. The request handler and other non-labeled portions of sd-ai are the generic infrastructure that future researchers need not implement if they choose to work within the sd-ai platform. Instead, the scope of the problem has been narrowed to the interaction with the underlying LLM alone.

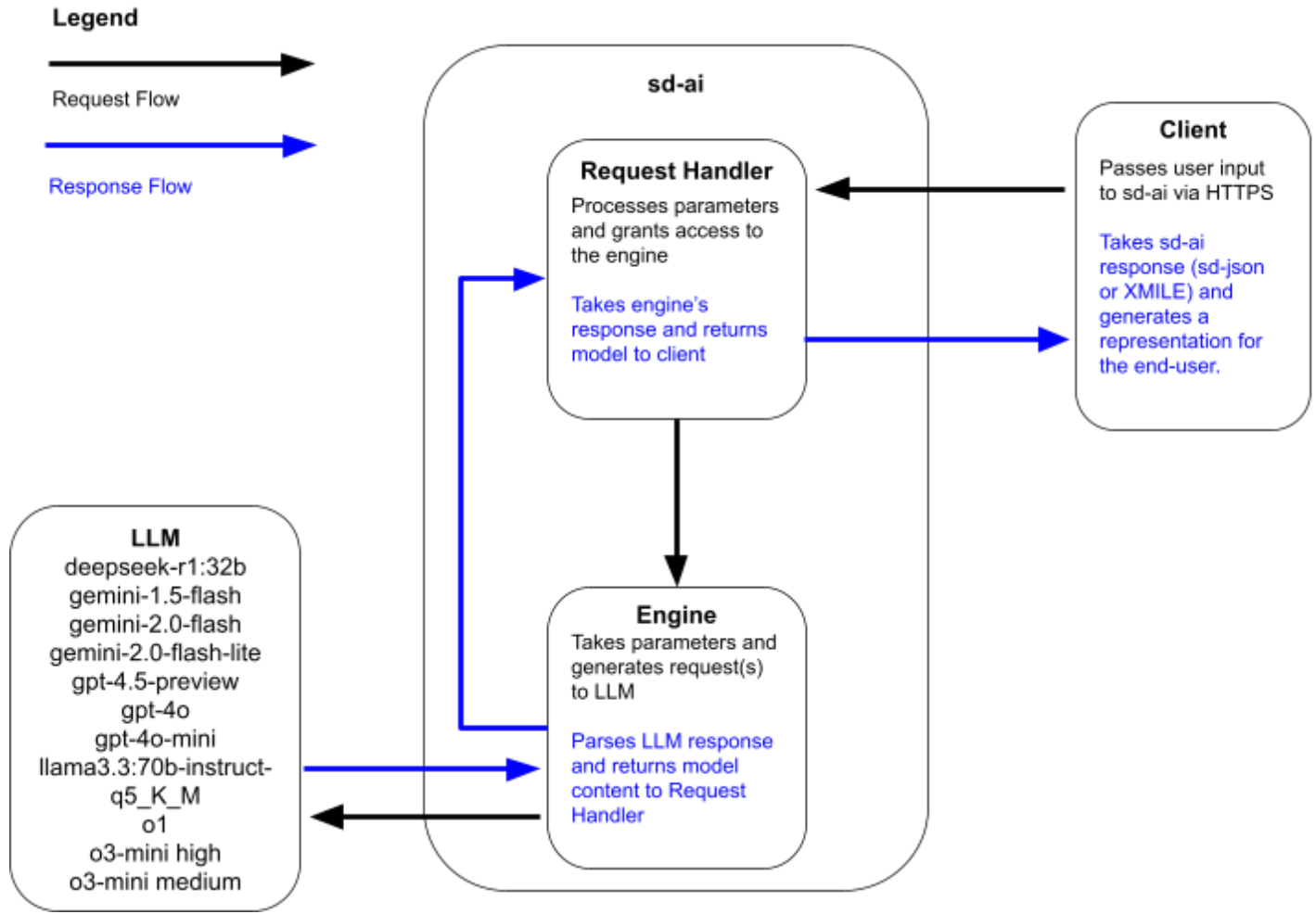

Figure 1: An architectural diagram showing how the sd-ai service handles requests from users, processes information, and returns model content back to the user.

An sd-ai engine is a javascript class which must implement two methods: the first is a function that generates the list of parameters on which the engine operates (i.e., the inputs it takes beyond a generic prompt from the end user, and the current state of the model); the second required method is the "generate" method that actually does the work of turning user-generated queries into models that take as arguments the user prompt, the current model, and all requested parameters. In the current release of sd-ai, there are 3 engines to choose from, though only two of which interact with LLMs. The first and simplest engine is 'predprey,' a dummy engine to demonstrate the concept to someone seeking to develop their own engine, which, when given any input, always returns a simple 2 variable, 2 relationship model showing predators reduce prey, and prey increase predators. The other two engines are the fully functional "advanced" engine and its "default" engine variant. The advanced engine exposes all of the options available for customization to the client, whereas the default engine reduces the full set of customization from the advanced engine down to only the most commonly used features, supplying default values instead. Because these two engines are identical under the default settings, we refer to both engines as "the sd-ai engine" in discussing their design and invoke the "default" or "advanced" qualifiers only when distinguishing between them. These two engines interact with a variety of LLMs as shown in Figure 1, and the functionality of the engines is described in the next section.

## 2.1 The sd-ai engine

The two LLM connected engines (advanced and default) included in the current release of sd-ai grew out of the research behind the SDBot project (Hosseinichimeh et al., 2024). The SDBot project leverages existing LLMs (specifically GPT 4 turbo) and uses a multi-shot, multi-pass prompting process to generate causal diagrams. The sd-ai engine builds upon that work, leveraging existing LLMs (to date without any customization or fine tuning), but instead uses a zero-shot, one-pass prompting scheme to generate models, taking advantage of efficiencies in information passing to the LLM to achieve at least comparable results.

To generate models using a zero-shot, one-pass prompt scheme, changes needed to be made to the prompt scheme employed by SDBot. For instance, the sd-ai engine makes use of the structured outputs (OpenAI Platform, 2025a) feature of current LLMs to guarantee parseable responses from the LLM (JSON); plus, structured outputs allow for a more efficient and directed set of instructions about the meaning of each element of the schema by which the engine communicates with the LLM to be communicated to the LLM. Making this change meant that the large majority of the SDBot system prompt (with the multiple example responses) was no longer required, including the part which describes the JSON schema by which SDBot and the LLM communicated. In addition to the efficiency gained in information passing via structured outputs, the use of structured outputs obviated the need for the second pass polarity prompt from SDBot, which was used to determine the polarity of each causal relationship in a secondary process after the relationship had been found. Finally removing the multi-shot aspect of the SDBot prompt scheme prevented information from the examples leaking into generated models.

The next significant change to the prompting scheme from SDBot was to include the ability to incrementally build models, supplying to the underlying LLM a pre-existing model to use as a starting point for a response. To support this functionality, the sd-ai engine makes use of the 'assistant' role (OpenAI Platform, 2025b) of modern LLMs to supply the stateless LLM with information about the model on which the user is currently working. This includes not only the full network of connections and polarities, but also any documentation or other reasoning embedded within that model that may provide additional information to the LLM for use in its own internal response generation process. Allowing the current state of an in-progress model to be shared with the LLM changes the role of the LLM in the model building process significantly as compared to the original SDBot implementation. This allows us to change the role of the LLM from that of an expert modeler who takes what the user sends and builds the entire model at once without additional user input (i.e., replaces the modeling process) to that of an assistant with whom that end user works to amend, improve, or otherwise develop their thinking incrementally. Our expectation is that this will make LLM to be more effective in providing suggestions overall by allowing more opportunity for the LLM to make pointed changes in response to specific instructions from the end user. Fundamentally, the improvement here is also about the efficiency of information delivery to the underlying LLM.

The next place where we made changes to the original SDBot prompting scheme was to more efficiently provide information about the problem being worked on and any background information that the end user may have access to that the LLM does not. By including input fields where the end user can specify a problem statement and include key background information, we enable the end user to structure their thinking (making tools backed by sd-ai work with existing modeling methodologies rather than attempting to replace them outright), while simultaneously providing relevant information that may not be in the original LLM's training set, such as interview transcripts, recent scientific publications, etc. By providing the underlying LLM with more information, an opportunity exists for the LLM to provide a higher quality response using that information.

The next set of changes as compared to SDBot involved the removal of the variable/relationship similarity detection schemes and the inclusion of stronger statements about the importance of presenting feedback in the returned model. The need for similarity detection and removal prompts were obviated by a combination of the switch to an incremental mode of interaction (ostensibly with clients that support end-user edits between subsequent requests to the engine) and the addition of structured outputs, which drastically reduced the incidence of duplicate relationship creation events. The need for the stronger statements about feedback was driven by a desire to reduce the incidence of diagram generation that included long causal chains without feedback into the rest of the system, which was a tendency we noticed during development.

Finally, for the advanced sd-ai engine only (i.e., default values are specified for these inputs in the default engine), to support researchers, especially those without a software engineering background who wish to get involved in the prompt engineering aspect of AI aided development, we externalized all prompt text embedded within the advanced sd-ai engine, as well as its underlying LLM selection, making it all end-user customizable. This allows clients to present to specific end users the opportunity to do their own prompt engineering using the prompting scheme of the advanced engine as well as to test using different LLMs to do model generation. The hope here is that this enables a wider audience of developers to get involved in the underlying technical research of interacting with LLMs to enhance system dynamics model creation.

## 2.2 Evaluating sd-ai engine outputs

Once we developed the sd-ai engine, it was important for us to start the process of evaluating the outputs of that engine or any other engine to be developed within the sd-ai framework. To produce a measure of quality requires breaking down the definition of quality (ultimately leading to utility, the full scope of which is beyond the present paper) into measurable criteria. To us, the first step in measuring quality is correctness. Correctness, aspects of which we strive to measure, is the ability of the engine to perform causal translation, i.e. distill relationships ("from" variable, "to" variable, and polarity) from plain (currently English) text. Another attribute of LLM performance that is relevant to measure to gain a sense of engine quality is conformance, i.e. the ability of the LLM to take instructions from the user on statements describing the desired output. We expect there are other attributes like these that, when measured as a group, would allow the user of an sd-ai engine to make judgements about its quality or usefulness, but for now the only two attributes for which we have developed rigorous and objective measurement criteria for are "causal translation" and "conformance".

### 2.2.1 Measuring causal translation abilities

To adequately and objectively measure what we define as causal translation (i.e., turning plain English text into structured data of causal relationships), it is necessary to create "fake alternate universes" in which we specify an objective synthetic ground truth. By shifting context into a gibberish world, we test the LLM's ability to extract causal relationships from provided data without relying on contextual clues from its own training data.

For each causal translation test, we need an algorithm to generate the ground truth in English that is fed to the LLM via the background knowledge prompt, and the corresponding graph network (model) that the written English represents from which to compare the output of the LLM. To build these ground truth networks and plain English descriptions, we created an algorithm to construct them starting with a list of 56 gibberish, non-pluralized nouns. We used those nouns (after a uniform pluralization process) as variable names and the basis for creating "causal sentences" that were strung together to form "causal descriptions" of the ground truth system. Causal sentences were built based on a "from" variable, a "to" variable, a polarity, and a polarity direction. The "from" variable is the cause, and the "to" variable is the effect variable. The polarity is either positive or negative, and the polarity direction is either up or down. A polarity direction of "up" means that the polarity in the causal sentence is described starting with an increase in the "from" variable, while "down" means a decrease in the "from" variable. Causal sentences are composed of the "from" variable, the "to" variable, and two adjectives describing the development of the "from" variable and the "to" variable. The form of causal sentences are "The [more|less] ["from" variable] there are, the [more|fewer] ["to" variable] there are." The "from" and "to" variables in causal sentences are always pluralized.

Our causal translation test suite contains 24 tests organized into three groups. The current test suite is designed to test the most basic form of causal translation (i.e., without the influence of confounders or more complicated potential forms of causal sentences). The first test group we've implemented is single relationship extraction, where there are 4 tests using a single causal sentence with gibberish words as variables. The full complement of polarity and polarity directions are tested here. The second group of causal translation tests consists of single loop extractions, where we measure the ability of the LLM to extract all of the relationships that compose a single feedback loop generated from causal sentences of gibberish words. Here we have 14 tests – 7 for each loop polarity (positive, negative) – where those 7 are feedback loops of variable length 2 to 7 inclusive for each polarity, and we alternate polarity directions for each link in each loop. The third group of tests are multiple loop extractions in which we measure the ability of the LLM to extract all of the relationships that compose a set of overlapping feedback loops of various lengths and polarities, where each feedback loop follows the same properties as the single loop extractions. Here we have two tests that assess the ability of the LLM to extract two overlapping loops of lengths 3 and 6 variables, two tests containing three overlapping loops each (containing 5, 2, and 4 variables, respectively), and the final two tests of five loops each that overlap each other where the loops contain 3, 5, 6, 2, and 6 variables respectively.

### 2.2.2 Measuring conformance to user requests

We measure conformance to end-user instruction on three primary attributes: 1) ability to include requested variables; 2) ability to adhere to instructions about the number of variables to include in the generated model; and 3) ability to adhere to instructions about the number of feedback loops to include in the generated model. To accurately capture the ability of current LLMs to follow directions of this sort, we must work with open-ended "real-world" contexts where there are multiple valid solutions that exist at different levels of complexity, each containing different variable names and different numbers of feedback loops for a given, fixed, problem statement and set of background knowledge. Context is necessary because conformance tests assess the ability of the LLM to create different representations at varying levels of complexity/aggregation for the same underlying system. If we were to give the LLM an alternate universe as the known ground truth, with a very specific set (essentially a relatively small number) of defined variables and relationships (as we do for causal translation), we wouldn't be able to test the ability of the LLM to properly simplify (or elaborate) an answer, because our generated alternate universes lack the flexibility of representation that real-world systems have. By definition, our alternate universe generated causal descriptions that have only one causal representation, so asking the LLM to simplify it or elaborate on it would be asking it to do something incorrect.

Therefore, to test each of these conformance attributes, we take two base prompts: the first that asks the LLMs to create a feedback-based explanation for the American revolutionary war, and a second that asks the LLMs to create a feedback-based explanation for road rage. We then append to that base each of the specific conformance commands:

1. Your response must include the variables
    a. (Revolution case) "Taxation", "Anti-British Sentiment" and "Colonial Identity"
    b. (Road rage case) "Traffic Congestion", "Driver Stress", and "Accidents"
2. Your response must include [at least|no more than] X variables.
    a. {at least 10 variables}
    b. {no more than 5 variables}
3. Your response must include [at least|no more than] X feedback loops.
    a. {at least 8 feedback loops}
    b. {no more than 4 feedback loops}
4. Your response must include [at least|no more than] X feedback loops and [at least|no more than] Y variables.
    a. {at least 6 feedback loops; at least 8 variables}
    b. {at least 6 feedback loops; no more than 15 variables}
    c. {no more than 4 feedback loops; no more than 5 variables}
    d. {no more than 4 feedback loops; at least 5 variables}

These instructions produce 9 tests for each case, so there are 18 total conformance tests. Values for X and Y are set to sufficiently challenge the LLMs over a range of specified conditions.

# 3. Results from testing the sd-ai engine with different LLMs

The results from testing the sd-ai engine using 11 different underlying LLMs on both causal translation and conformance attributes are presented in Tables 1-3 below. Table 1 shows high level performance on each of the two test suites individually and overall. The highest performing LLM overall is OpenAI's gpt-4.5-preview, which passed 92.9% of the tests overall and performed the best of any LLM examined on both causal translation and conformance. gpt-4.5-preview was one of two LLMs that scored 100% on causal translation; the second was o1. The next two highest performers on causal translation were the o3 family of LLMs, where o3-mini with high reasoning scored 87.5% (3 failures, all fake relationships) and o3-mini with medium reasoning scored 91.7% (2 failures, one fake relationship, one with multiple failures involving a reversal of the "from" and "to" variables of a relationship). Examining the causal translation failure modes in Table 2 shows that an interesting performer was gpt-4o, which scored relatively poorly on causal reasoning overall (41.7%), but when digging deeper into those failures (see Appendix 1), the only causal translation failure (repeated 14 times) was mis-labeling the polarity of reinforcing relationships when the corresponding causal sentence was expressed in terms of decreasing words (the less X, the fewer Y).

On conformance testing, gpt-4.5-preview was the highest performer, getting a score of 83.3% with only 3 failures, as indicated in Table 3. gpt-4o was second, scoring 77.8% with only 4 failures, and third place was a tie between o1, gpt-4o-mini, and llama3.3:70b-instruct-q5_K_M with 5 failures and a score of 72.2%. Failures on conformance tests were generally due to a failure to include feedback even in the highest performing LLMs. No LLM failed to create variables when instructed to include them specifically; of the 3 instances of multiple failures in conformance testing, none of them were due in part to this test (see Appendix 1 for detail on multiple failures). Of the three LLMs (gpt-4.5-preview, o1, and gpt-4o) that were able to fully recover the causal network (i.e., to not generate fake relationships or omit real relationships), they all performed similarly on conformance, with their failures being heavily concentrated in not returning enough feedback. Table 1 shows that LLMs are marginally better at conforming to user instruction than causal translation, with the average conformance score being 66.7%, and the average causal translation score being 49.6%.

Table 1: Overall results of LLM performance on causal translation and conformance tasks. Percentages indicate the proportion of 24 causal translation tests, 18 conformance tests, and 42 total tests that were executed successfully.

| LLM | Causal Translation | Conformance | Overall |
|---|---|---|---|
| deepseek-r1:32b | 16.7% | 44.4% | **28.6%** |
| gemini-1.5-flash | 50.0% | 61.1% | **54.8%** |
| gemini-2.0-flash | 50.0% | 55.6% | **52.4%** |
| gemini-2.0-flash-lite | 25.0% | 61.1% | **40.5%** |
| gpt-4.5-preview | 100.0% | 83.3% | **92.9%** |
| gpt-4o | 41.7% | 77.8% | **57.1%** |
| gpt-4o-mini | 4.2% | 72.2% | **33.3%** |
| llama3.3:70b-instruct-q5_K_M | 20.8% | 72.2% | **42.9%** |
| o1 | 100.0% | 72.2% | **88.1%** |
| o3-mini high | 87.5% | 66.7% | **78.6%** |
| o3-mini medium | 91.7% | 66.7% | **81.0%** |

Table 2: Explanation of causal translation failures out of 24 tests. If a single test failed both because a real relationship was not found, and a relationship had the wrong polarity, then the failure was counted as "Multiple Kinds of Failure".

| | **Causal Translation Testing Failure Reason** | | | | |
|---|---|---|---|---|---|
| **Reason for Failure** | **Fake relationship** | **Missing relationship** | **Multiple Kinds of Failures** | **Polarity** | **Grand Total** |
| deepseek-r1:32b | 1 | 2 | 7 | 10 | 20 |
| gemini-1.5-flash | 2 | | 1 | 9 | 12 |
| gemini-2.0-flash | 5 | | 2 | 5 | 12 |
| gemini-2.0-flash-lite | 1 | | 9 | 8 | 18 |
| gpt-4.5-preview | | | | | 0 |
| gpt-4o | | | | 14 | 14 |
| gpt-4o-mini | 2 | | 13 | 8 | 23 |
| llama3.3:70b-instruct-q5_K_M | 2 | | 11 | 6 | 19 |
| o1 | | | | | 0 |
| o3-mini high | 3 | | | | 3 |
| o3-mini medium | 1 | | 1 | | 2 |

Table 3: Explanation of conformance failures out of 18 tests. If a single test failed both because the LLM failed to both adhere to a variable and feedback constraint simultaneously, then the failure was counted as "Multiple Kinds of Failure".

|  | Conformance Testing Failure Reason | | | | | |
| --- | --- | --- | --- | --- | --- | --- |
| Reason for Failure | Multiple Kinds of Failures | Too few feedback loops | Too few variables | Too many feedback loops | Too many variables | Grand Total |
| deepseek-r1:32b |  | 5 | 2 |  | 3 | 10 |
| gemini-1.5-flash |  | 5 |  |  | 2 | 7 |
| gemini-2.0-flash | 1 | 3 |  | 1 | 3 | 8 |
| gemini-2.0-flash-lite | 2 | 2 | 1 | 1 | 1 | 7 |
| gpt-4.5-preview |  | 2 | 1 |  |  | 3 |
| gpt-4o |  | 3 |  | 1 |  | 4 |
| gpt-4o-mini |  | 3 | 2 |  |  | 5 |
| llama3.3:70b-instruct-q5_K_M |  | 2 |  | 2 | 1 | 5 |
| o1 |  | 4 |  | 1 |  | 5 |
| o3-mini high |  | 6 |  |  |  | 6 |
| o3-mini medium |  | 5 |  | 1 |  | 6 |

These results indicate that the most technically correct LLMs tested were gpt-4.5-preview and o1, though an interesting result is that gpt-4o, which is the oldest of the three LLMs that were capable of correctly identifying all causal links, is also the cheapest, and is significantly faster to run (Artificial Analysis, 2025). Its only fault relative to the two newer and far more expensive LLMs was interpreting positive polarity when specified in decreasing terms. In general, these results demonstrate that sd-ai users who use gpt-4.5-preview, o1, and gpt-4o should expect high quality causal translations but generally simpler models than requested. In practice, users can combat the simplification tendency of these LLMs by requesting more variable names as a part of their prompting, trusting in some sense the causal translation ability of the LLMs to fill in the relationships between the variables they request.

# 4. Responsibly developing AI tools for model building

The results of the causal translation and conformance tests articulated above demonstrate the feasibility of measuring the ability of the sd-ai engine to produce causal maps with different LLMs. The attribute of causal translation reflects the accuracy and completeness of the causal relationships and feedback mechanisms generated by an AI-enabled tool for scenarios in which such causal relationships are fully specified. The conformance attribute reflects the ability of the AI tool to produce a reasonable model structure from the background knowledge provided and following the instructions given. Tests for correctness and ability to follow instructions are necessary but not sufficient to determine how well AI-enabled tools can be used for model building. These straightforward tests nevertheless provide an extensible foundation for assessing the adequacy of AI-enabled tools for dynamic modeling. From this foundation, new tests will be added to the sd-ai project to measure additional aspects of AI tools that in turn incentivize responsible AI tool development to improve model building. Additional tests will be needed to assess the quality of model structures generated by AI tools and how useful these tools are for solving problems. Effectively prioritizing and developing tests for assessing the ability of AI tools to perform modeling tasks will require thoughtful human collaboration.

The sd-ai project represents a collaborative effort to integrate AI into the modeling process, beginning with the construction of causal maps. Collaboration was needed not just for the design of the sd-ai engine but also for the identification of attributes by which to measure its performance using different LLMs. Based on this experience, we contend that harnessing human intelligence through collaboration will be essential to guide future development of AI tools that can support and enhance the modeling process. To engage the wider AI community in this effort, we have launched the Benchmarking and Evaluating AI for Modeling and Simulation (BEAMS) Initiative as a collaboration among tool developers, modelers, and stakeholders. BEAMS is an official research project of the Institute for Artificial Intelligence and Data Science at the University at Buffalo.

The rapid pace of AI development underscores the need we have as humans to adequately evaluate the ability of AI-enabled tools to perform modeling tasks. The BEAMS Initiative thus aims to motivate responsible and ethical AI tools for model building through a process of continued evaluation and quality improvement. This is broadly a three-step process of asking and answering the following questions:

> Step 1: How can AI-based automation improve dynamic modeling?
> Step 2: What aspects of AI-based tools and their outputs can be measured and evaluated to spur their development in a way that improves dynamic modeling?
> Step 3: How do we measure those aspects?

As depicted in Figure 2 below, this collaborative evaluation process is designed to occur through two working groups of the BEAMS Initiative: a steering group to prioritize how AI-enabled tools for modeling should work (Step 1) and a technical group to implement specific measures of how well they work (Step 3). The scope of these groups overlaps with Step 2, so the steering and technical groups will work together to identify what aspects of AI tools can be measured to reflect the desired specification from Step 1.

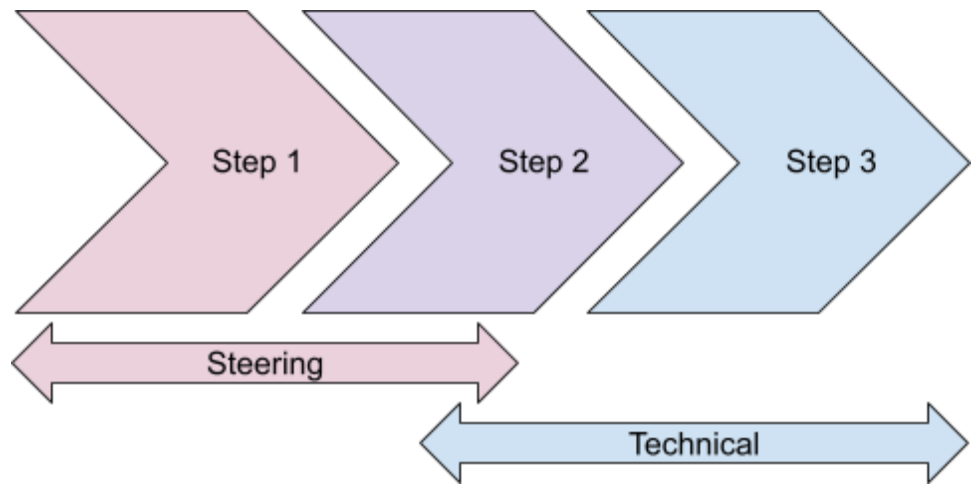

Figure 2: Overview of the three-step collaborative process that the steering and technical groups of the BEAMS Initiative will follow to prioritize and implement benchmarks for how well AI tools can build dynamic models.

While this three-step process charts a general progression from determining how AI can improve modeling (Step 1) to how we can tell (Step 3), collaboration in the BEAMS Initiative will also occur in an iterative manner. This process includes reflecting back what was defined in Step 2 relative to what was specified in Step 1, and reporting back on what was implemented in Step 3 based on what was defined in Step 2. Such feedback will provide opportunities for further refinement of measurable benchmarks and consideration of new avenues of inquiry that may have been opened up along the way. For instance, a potential refinement of tests for causal translation would be to include confounding factors when assessing the ability of AI-enabled tools to identify cause and effect relationships.

The collaborative process of benchmarking and evaluating AI tools as envisioned for the BEAMS Initiative can be illustrated by insights from the efforts of the sd-ai project thus far. If the ability of an AI tool to construct high-quality causal maps is identified as a priority by the steering group in Step 1, then the steering and technical groups will work in tandem on Step 2 to identify what aspects of the tool can be measured as a barometer for causal map quality, such as the ability of an AI tool to perform causal reasoning. The technical group will then execute the implementation of specific measures for causal reasoning in Step 3, including but not limited to the causal translation and conformance tests described earlier. Both groups would then revisit Step 2 to consider whether these tests sufficiently capture the causal reasoning capacity of an AI tool.

Another aspect of quality in a causal map is its representation of appropriate concepts. To measure this aspect of quality in causal reasoning, tests for conceptual overlap may be developed in which correspondence between variables in the AI-generated causal map and the expected constructs are assessed using a different LLM from the one that was invoked to create the causal map. While the conformance tests included instructions to require certain variables verbatim, conceptual overlap tests would assess whether specified concepts are included in the causal map. Such tests would allow for the complexity of the real world to be incorporated without defining aspects of it, neither reference-based nor reference-free in the taxonomy of Giabbanelli et al. (2025).

Importantly, the aspects of AI-based tools that can be assessed through this collaborative evaluation framework are not just their outputs, but also their assistance in the iterative process of model development. In this vein, a novel set of tests may be defined to evaluate the ability of an AI tool to build on given input and thereby enhance the modeling process. As noted above, the sd-ai engine was designed to enable incremental model building, so that a pre-existing model could be provided as a starting point for further development. We expect that evaluating a tool's ability to interact with user input will incentivize development of AI tools that assist rather than replace human modelers.

The emphasis of the BEAMS Initiative is on responsibly developing AI tools for modeling and simulation writ large, extending beyond the scope of systems thinking and system dynamics per se. Measurements of causal translation and conformance that are applied to causal maps produced by AI tools provide a powerful starting point for this process. Going forward, in addition to devising further tests of causal map construction, the BEAMS Initiative will develop tests for the ability of AI tools to generate other model structures, such as stocks and flows, and to specify appropriate parameters for such structures to be simulation-ready. Moreover, this collaborative evaluation process will articulate tests of how well an AI tool can generate algorithms and interactions for agent-based modeling. Giabbanelli (2023) describes different ways of integrating AI into simulation tasks and underscores the accessibility benefit of harnessing AI for interpretation of model structures in narrative form.

Ultimately, the BEAMS Initiative and the open source sd-ai project provide a framework for generating more comprehensive tests, motivating better LLMs and sd-ai engines that can not just make better models but also can make us better modelers. This framework relies on collaboration to foster AI tools that can deepen human understanding and expand access to insights from dynamic modeling. Human guidance is needed for development of this technology to ensure it improves the modeling process for humans. Absent thoughtful guidance, technology will tend toward the path of least resistance rather than optimal utilization.

# 5. References

Artificial Analysis. (2025). Independent analysis of AI models and API providers. Retrieved March 9, 2025 from https://artificialanalysis.ai/

Barbrook-Johnson, P., Penn, A.S. (2022). Causal Loop Diagrams. In: *Systems Mapping.* Palgrave Macmillan, Cham. https://doi.org/10.1007/978-3-031-01919-7_4

Barlas, Y. (1996). Formal aspects of model validity and validation in system dynamics. *System Dynamics Review, 12* (3), 183-210.

Black, L. & Greer, D. (2024). Minding the abstraction gap: approaches supporting implementation. *System Dynamics Review, 40* (4), e1790.

Brooks, R. (1986). A robust layered control system for a mobile robot. *IEEE Journal of Robotics and Automation*, *2* (1), 14–23.

Chaniotis, I. K., Kyriakou, K. I. D., & Tselikas, N. D. (2015). Is Node. js a viable option for building modern web applications? A performance evaluation study. *Computing*, *97*, 1023-1044.

du Plooy, C., & Oosthuizen, R. (2023). AI usefulness in systems modelling and simulation: GPT-4 application. *South African Journal of Industrial Engineering*, *34*(3), 286-303. https://doi.org/10.7166/34-3-2944

Forrester J. W. (1961). *Industrial Dynamics.* Pegasus Communications: Waltham, MA.

Forrester, J. W. (1980). Information sources for modeling the national economy. *Journal of the American Statistical Association, 75* (371), 555-574.

Forrester, J. W. (2007). System dynamics—a personal view of the first fifty years. *System Dynamics Review*, *23* (2‐3), 345-358.

Giabbanelli, P. J. (2023). GPT-based models meet simulation: How to efficiently use large-scale pre-trained language models across simulation tasks. In *Proceedings of the Winter Simulation Conference* (pp. 2920-2931). https://arxiv.org/pdf/2306.13679

Giabbanelli, P. J., Gandee, T. J., Agrawal, A., & Hosseinichimeh, N. (2025). Benchmarking and assessing transformations between text and causal maps via Large Language Models. *Applied Ontology,* https://journals.sagepub.com/doi/pdf/10.1177/15705838241304102

Hirsch-Kreinsen, H. (2024). Artificial intelligence: a "promising technology". *AI & Soc 39*, 1641–1652. https://doi.org/10.1007/s00146-023-01629-w

Homer, J. (2013). The aimless plateau, revisited: why the field of system dynamics needs to establish a more coherent identity. *System Dynamics Review, 29* (2), 124-127.

Hu, B. (2025). CHATPYSD: Embedding and Simulating System Dynamics Models in CHATGPT -4. *System Dynamics Review*, *41*(1), e1797. https://doi.org/10.1002/sdr.1797

Hosseinichimeh, N., Majumdar, A., Williams, R., & Ghaffarzadegan, N. (2024). From text to map: a system dynamics bot for constructing causal loop diagrams. *System Dynamics Review*, 40(3), e1782.

Jalali, M. S. & Akhavan, A. (2024). Integrating AI language models in qualitative research: Replicating interview data analysis with ChatGPT. *System Dynamics Review*, 40(3), e1772.

Liu, N. Y. G. & Keith, D. (2024). Leveraging large language models for automated causal loop diagram generation: enhancing system dynamics modeling through curated prompting techniques. *SSRN*. https://ssrn.com/abstract=4906094

Martin Jr, D., Prabhakaran, V., Kuhlberg, J., Smart, A., & Isaac, W. S. (2020). Participatory problem formulation for fairer machine learning through community based system dynamics. *arXiv preprint arXiv:2005.07572*.

Monteith, S., Glenn, T., Geddes, J. R., Whybrow, P. C., Achtyes, E., & Bauer, M. (2024). Artificial intelligence and increasing misinformation. *British Journal of Psychiatry, 224*(2), 33–35. https://doi.org/10.1192/bjp.2023.136

Neumann, A., Laranjeiro, N., & Bernardino, J. (2018). An analysis of public REST web service APIs. IEEE Transactions on Services Computing, 14(4), 957-970.

OpenAI Platform. (2025a). Structured Outputs. Retrieved March 4, 2025 from https://platform.openai.com/docs/guides/structured-outputs

OpenAI Platform. (2025b). Text Generation. Retrieved March 4, 2025 from https://platform.openai.com/docs/guides/text-generation

Pruyt, E., Cunningham, S., Kwakkel, J. H., & de Bruijn, J. A. (2014). From data-poor to data-rich: system dynamics in the era of big data. In *Proceedings of the 32nd International Conference of the System Dynamics Society* (pp. 2458-2469)

Rahmandad, H. & Sterman, J. (2012). Reporting guidelines for simulation-based research in social sciences. *System Dynamics Review, 28* (4), 396-411.

Richardson, G. (1999). Reflections for the future of system dynamics. *Journal of the Operational Research Society, 50*(4), 440-449. DOI: 10.1057/palgrave.jors.2600749

Salatino, A., Osborne, F., & Motta, E. (2024). Artificial intelligence for literature reviews: Opportunities and challenges. *Artificial Intelligence Review*. SpringerLink

Schwaninger, M. (2019). On John Sterman's "System dynamics at sixty": rigor, relevance and implications for education. *System Dynamics Review, 35 (*1), 15-18.

Senge, P. M., & Forrester, J. W. (1980). Tests for building confidence in system dynamics models. *System Dynamics, TIMS Studies in Management Sciences*, *14* (14), 209-228.

Shin, D. (2024). Misinformation and Algorithmic Bias. In: *Artificial Misinformation*. Palgrave Macmillan, Cham. https://doi.org/10.1007/978-3-031-52569-8_2

Simon, H. A. (1969). *The sciences of the artificial*. MIT Press.

Sterman, J. (2018). System dynamics at sixty: the path forward. *System Dynamics Review, 34* (1-2), 5-47.

Veldhuis, G.A., Blok, D., de Boer, M. H.T., Kalkman, G. J., Bakker, R. M., & van Waas, R. P. M. (2024). From test to model: Leveraging natural language processing for system dynamics model development. *System Dynamics Review, 40* (3), 1-26.

Weidinger, L., Uesato, J., Maribeth Rauh, M., Griffin, C., Huang, P.-S., Mellor, J., Glaese, A., Cheng, M., Balle, B., Kasirzadeh, A., Biles, C., Brown, S., Kenton, Z., Hawkins, W., Stepleton, T., Birhane, A., Hendricks, L. A., Rimell, L., Isaac, W., Haas, J., Legassick, S., Irving, G., & Gabriel, I. (2022). Taxonomy of Risks posed by Language Models. In Proceedings of the 2022 ACM Conference on Fairness, Accountability, and Transparency (FAccT '22). Association for Computing Machinery, New York, NY, USA, 214–229. https://doi.org/10.1145/3531146.3533088

Xu, Y., Wang, F., & Zhang, T. (2024). Artificial intelligence is restructuring a new world. *Innovation (Cambridge (Mass.))*, *5* (6), 100725. https://doi.org/10.1016/j.xinn.2024.100725

# Appendix 1 - Log of all test failures

Failures:
1) gemini-2.0-flash-lite | causal translation testing | single feedback loop | extract a reinforcing feedback loop with 2 variables
  Message:
    Incorrect polarity discovered: Expected '-' to be '+'.

2) gemini-2.0-flash-lite | causal translation testing | multiple feedback loops | extract 3 feedback loops with [+, +, -] polarities
  Message:
    Incorrect polarity discovered: Expected '-' to be '+'.

  Message:
    Incorrect polarity discovered: Expected '-' to be '+'.

  Message:
    Incorrect polarity discovered: Expected '-' to be '+'.

  Message:
    Incorrect polarity discovered: Expected '-' to be '+'.

  Message:
    Incorrect polarity discovered: Expected '-' to be '+'.

3) gemini-2.0-flash-lite | causal translation testing | single feedback loop | extract a reinforcing feedback loop with 5 variables
  Message:
    Incorrect polarity discovered: Expected '-' to be '+'.

  Message:
    Incorrect polarity discovered: Expected '-' to be '+'.

  Message:
    Incorrect polarity discovered: Expected '-' to be '+'.

4) gemini-2.0-flash-lite | causal translation testing | single relationship | extract a reinforcing relationship down
  Message:
    Fake relationships found
    Fewer Frimbulators --> (+) Fewer Whatajigs
    Ground Truth
    frimbulators --> (+) whatajigs: Expected 1 to be 0.

  Message:
    Real relationships not found

frimbulators --> (+) whatajigs
    Ground Truth
    frimbulators --> (+) whatajigs: Expected 1 to be 0.

5) gemini-2.0-flash-lite | causal translation testing | single feedback loop | extract a reinforcing feedback loop with 7 variables
   Message:
    Incorrect polarity discovered: Expected '-' to be '+'.

   Message:
    Incorrect polarity discovered: Expected '-' to be '+'.

   Message:
    Incorrect polarity discovered: Expected '-' to be '+'.

   Message:
    Incorrect polarity discovered: Expected '-' to be '+'.

6) gemini-2.0-flash-lite | causal translation testing | single feedback loop | extract a reinforcing feedback loop with 4 variables
   Message:
    Fake relationships found
    funkados --> (-) refluppers, marticatenes --> (+) maxabizers, maxabizers --> (+) funkados, refluppers --> (+) marticatenes
    Ground Truth
    funkados --> (+) maxabizers, marticatenes --> (+) refluppers, maxabizers --> (+) marticatenes, refluppers --> (+) funkados: Expected 4 to be 0.

   Message:
    Incorrect polarity discovered: Expected '-' to be '+'.

   Message:
    Incorrect polarity discovered: Expected '-' to be '+'.

7) gemini-2.0-flash-lite | causal translation testing | single feedback loop | extract a reinforcing feedback loop with 6 variables
   Message:
    Incorrect polarity discovered: Expected '-' to be '+'.

   Message:
    Incorrect polarity discovered: Expected '-' to be '+'.

   Message:
    Incorrect polarity discovered: Expected '-' to be '+'.

8) gemini-2.0-flash-lite | causal translation testing | multiple feedback loops | extract 2 feedback loops with [-, +] polarities
  Message:
    Fake relationships found
    refluppers --> (-) frimbulators
    Ground Truth
    balacks --> (+) frimbulators, balacks --> (+) whoziewhats, frimbulators --> (-) whatajigs, funkados --> (+) maxabizers, marticatenes --> (+) refluppers, maxabizers --> (+) marticatenes, refluppers --> (+) balacks, whatajigs --> (+) balacks, whoziewhats --> (+) funkados: Expected 1 to be 0.

9) gemini-2.0-flash-lite | causal translation testing | multiple feedback loops | extract 2 feedback loops with [+, +] polarities
  Message:
    Fake relationships found
    frimbulators --> (+) maxabizers, frimbulators --> (+) refluppers, frimbulators --> (+) whoziewhats, frimbulators --> (-) funkados, frimbulators --> (-) marticatenes, refluppers --> (+) frimbulators
    Ground Truth
    balacks --> (+) frimbulators, balacks --> (+) whoziewhats, frimbulators --> (+) whatajigs, funkados --> (+) maxabizers, marticatenes --> (+) refluppers, maxabizers --> (+) marticatenes, refluppers --> (+) balacks, whatajigs --> (+) balacks, whoziewhats --> (+) funkados: Expected 6 to be 0.

  Message:
    Incorrect polarity discovered: Expected '-' to be '+'.

  Message:
    Incorrect polarity discovered: Expected '-' to be '+'.

  Message:
    Incorrect polarity discovered: Expected '-' to be '+'.

  Message:
    Incorrect polarity discovered: Expected '-' to be '+'.

  Message:
    Incorrect polarity discovered: Expected '-' to be '+'.

10) gemini-2.0-flash-lite | causal translation testing | single feedback loop | extract a reinforcing feedback loop with 8 variables
  Message:
    Fake relationships found
    dominitoxings --> (-) ocs, ocs --> (-) priaries, povals --> (-) exemintes, priaries --> (-) povals
    Ground Truth
    auspongs --> (+) dominitoxings, dominitoxings --> (+) exemintes, exemintes --> (+) ocs, houtals --> (+) povals, ocs --> (+) proptimatires, povals --> (+) auspongs, priaries --> (+) houtals, proptimatires --> (+) priaries: Expected 4 to be 0.

Message:
  Incorrect polarity discovered: Expected '-' to be '+'.

Message:
  Incorrect polarity discovered: Expected '-' to be '+'.

Message:
  Incorrect polarity discovered: Expected '-' to be '+'.

Message:
  Incorrect polarity discovered: Expected '-' to be '+'.

11) gemini-2.0-flash-lite | causal translation testing | multiple feedback loops | extract 5 feedback loops with [-, +, +, -, -] polarities
  Message:
    Incorrect polarity discovered: Expected '-' to be '+'.

  Message:
    Incorrect polarity discovered: Expected '-' to be '+'.

  Message:
    Incorrect polarity discovered: Expected '-' to be '+'.

  Message:
    Incorrect polarity discovered: Expected '-' to be '+'.

  Message:
    Incorrect polarity discovered: Expected '-' to be '+'.

  Message:
    Incorrect polarity discovered: Expected '-' to be '+'.

  Message:
    Incorrect polarity discovered: Expected '-' to be '+'.

  Message:
    Incorrect polarity discovered: Expected '-' to be '+'.

  Message:
    Incorrect polarity discovered: Expected '-' to be '+'.

12) gemini-2.0-flash-lite | causal translation testing | single feedback loop | extract a balancing feedback loop with 8 variables
  Message:
    Incorrect polarity discovered: Expected '-' to be '+'.

Message:
    Incorrect polarity discovered: Expected '-' to be '+'.

  Message:
    Incorrect polarity discovered: Expected '-' to be '+'.

13) gemini-2.0-flash-lite | causal translation testing | single feedback loop | extract a balancing feedback loop with 4 variables
  Message:
    Fake relationships found
    Funkados --> (-) Marticatenes, Maxabizers --> (+) Refluppers
    Ground Truth
    funkados --> (-) maxabizers, marticatenes --> (+) refluppers, maxabizers --> (+) marticatenes, refluppers --> (+) funkados: Expected 2 to be 0.

  Message:
    Incorrect polarity discovered: Expected '-' to be '+'.

14) gemini-2.0-flash-lite | causal translation testing | multiple feedback loops | extract 5 feedback loops with [-, +, +, +, -] polarities
  Message:
    Incorrect polarity discovered: Expected '-' to be '+'.

  Message:
    Incorrect polarity discovered: Expected '-' to be '+'.

  Message:
    Incorrect polarity discovered: Expected '-' to be '+'.

  Message:
    Incorrect polarity discovered: Expected '-' to be '+'.

  Message:
    Incorrect polarity discovered: Expected '-' to be '+'.

  Message:
    Incorrect polarity discovered: Expected '-' to be '+'.

  Message:
    Incorrect polarity discovered: Expected '-' to be '+'.

  Message:
    Incorrect polarity discovered: Expected '-' to be '+'.

  Message:

    Incorrect polarity discovered: Expected '-' to be '+'.

  Message:
    Incorrect polarity discovered: Expected '-' to be '+'.

15) gemini-2.0-flash-lite | causal translation testing | single feedback loop | extract a balancing feedback loop with 7 variables
  Message:
    Fake relationships found
    exemintes --> (-) proptimatires, houtals --> (+) refluppers, ocs --> (+) priaries, priaries --> (-) povals, proptimatires --> (+) houtals, refluppers --> (+) ocs
    Ground Truth
    exemintes --> (+) ocs, houtals --> (+) povals, ocs --> (+) proptimatires, povals --> (+) refluppers, priaries --> (+) houtals, proptimatires --> (+) priaries, refluppers --> (-) exemintes: Expected 6 to be 0.

  Message:
    Incorrect polarity discovered: Expected '-' to be '+'.

  Message:
    Incorrect polarity discovered: Expected '-' to be '+'.

  Message:
    Incorrect polarity discovered: Expected '-' to be '+'.

16) gemini-2.0-flash-lite | causal translation testing | single feedback loop | extract a balancing feedback loop with 3 variables
  Message:
    Fake relationships found
    maxabizers --> (-) funkados
    Ground Truth
    funkados --> (+) maxabizers, maxabizers --> (+) whoziewhats, whoziewhats --> (-) funkados: Expected 1 to be 0.

  Message:
    Incorrect polarity discovered: Expected '-' to be '+'.

17) gemini-2.0-flash-lite | causal translation testing | single feedback loop | extract a balancing feedback loop with 6 variables
  Message:
    Fake relationships found
    exemintes --> (-) proptimatires, marticatenes --> (-) exemintes, ocs --> (+) priaries, priaries --> (+) refluppers, proptimatires --> (-) marticatenes, refluppers --> (+) ocs
    Ground Truth
    exemintes --> (+) ocs, marticatenes --> (-) refluppers, ocs --> (+) proptimatires, priaries --> (+) marticatenes, proptimatires --> (+) priaries, refluppers --> (+) exemintes: Expected 6 to be 0.

Message:
　　Incorrect polarity discovered: Expected '-' to be '+'.

　　Message:
　　Incorrect polarity discovered: Expected '-' to be '+'.

18) gemini-2.0-flash-lite | causal translation testing | single feedback loop | extract a reinforcing feedback loop with 3 variables
　　Message:
　　Fake relationships found
　　fewer funkados --> (-) less whoziewhats, fewer whoziewhats --> (-) less maxabizers, less maxabizers --> (-) fewer whoziewhats, less whoziewhats --> (-) fewer funkados, more funkados --> (+) more maxabizers, more maxabizers --> (+) more funkados
　　Ground Truth
　　funkados --> (+) maxabizers, maxabizers --> (+) whoziewhats, whoziewhats --> (+) funkados: Expected 6 to be 0.

　　Message:
　　Real relationships not found
　　funkados --> (+) maxabizers, maxabizers --> (+) whoziewhats, whoziewhats --> (+) funkados
　　Ground Truth
　　funkados --> (+) maxabizers, maxabizers --> (+) whoziewhats, whoziewhats --> (+) funkados: Expected 3 to be 0.

19) gemini-1.5-flash | causal translation testing | single feedback loop | extract a balancing feedback loop with 6 variables
　　Message:
　　Incorrect polarity discovered: Expected '-' to be '+'.

　　Message:
　　Incorrect polarity discovered: Expected '-' to be '+'.

20) gemini-1.5-flash | causal translation testing | single feedback loop | extract a reinforcing feedback loop with 5 variables
　　Message:
　　Incorrect polarity discovered: Expected '-' to be '+'.

　　Message:
　　Incorrect polarity discovered: Expected '-' to be '+'.

　　Message:
　　Incorrect polarity discovered: Expected '-' to be '+'.

21) gemini-1.5-flash | causal translation testing | single feedback loop | extract a balancing feedback loop with 5 variables
  Message:
    Incorrect polarity discovered: Expected '-' to be '+'.

  Message:
    Incorrect polarity discovered: Expected '-' to be '+'.

22) gemini-1.5-flash | causal translation testing | single feedback loop | extract a reinforcing feedback loop with 2 variables
  Message:
    Incorrect polarity discovered: Expected '-' to be '+'.

23) gemini-1.5-flash | causal translation testing | multiple feedback loops | extract 2 feedback loops with [-, +] polarities
  Message:
    Incorrect polarity discovered: Expected '-' to be '+'.

  Message:
    Incorrect polarity discovered: Expected '-' to be '+'.

  Message:
    Incorrect polarity discovered: Expected '-' to be '+'.

  Message:
    Incorrect polarity discovered: Expected '-' to be '+'.

24) gemini-1.5-flash | causal translation testing | multiple feedback loops | extract 2 feedback loops with [+, +] polarities
  Message:
    Fake relationships found
    balacks --> (-) refluppers, balacks --> (-) whatajigs, whatajigs --> (-) frimbulators
    Ground Truth
    balacks --> (+) frimbulators, balacks --> (+) whoziewhats, frimbulators --> (+) whatajigs, funkados --> (+) maxabizers, marticatenes --> (+) refluppers, maxabizers --> (+) marticatenes, refluppers --> (+) balacks, whatajigs --> (+) balacks, whoziewhats --> (+) funkados: Expected 3 to be 0.

25) gemini-1.5-flash | causal translation testing | single feedback loop | extract a balancing feedback loop with 7 variables
  Message:
    Incorrect polarity discovered: Expected '-' to be '+'.

  Message:
    Incorrect polarity discovered: Expected '-' to be '+'.

  Message:
    Incorrect polarity discovered: Expected '-' to be '+'.

26) gemini-1.5-flash | causal translation testing | single relationship | extract a balancing relationship up
  Message:
    Fake relationships found
    whatajigs --> (+) frimbulators
    Ground Truth
    frimbulators --> (-) whatajigs: Expected 1 to be 0.

27) gemini-1.5-flash | causal translation testing | single feedback loop | extract a reinforcing feedback loop with 7 variables
  Message:
    Incorrect polarity discovered: Expected '-' to be '+'.

  Message:
    Incorrect polarity discovered: Expected '-' to be '+'.

  Message:
    Incorrect polarity discovered: Expected '-' to be '+'.

  Message:
    Incorrect polarity discovered: Expected '-' to be '+'.

28) gemini-1.5-flash | causal translation testing | multiple feedback loops | extract 5 feedback loops with [-, +, +, +, -] polarities
  Message:
    Incorrect polarity discovered: Expected '-' to be '+'.

  Message:
    Incorrect polarity discovered: Expected '-' to be '+'.

  Message:
    Incorrect polarity discovered: Expected '-' to be '+'.

  Message:
    Incorrect polarity discovered: Expected '-' to be '+'.

  Message:
    Incorrect polarity discovered: Expected '-' to be '+'.

  Message:
    Incorrect polarity discovered: Expected '-' to be '+'.

Message:
    Incorrect polarity discovered: Expected '-' to be '+'.

  Message:
    Incorrect polarity discovered: Expected '-' to be '+'.

  Message:
    Incorrect polarity discovered: Expected '-' to be '+'.

  Message:
    Incorrect polarity discovered: Expected '-' to be '+'.

29) gemini-1.5-flash | causal translation testing | multiple feedback loops | extract 3 feedback loops with [-, -, +] polarities
  Message:
    Real relationships not found
    maxabizers --> (+) funkados
    Ground Truth
    balacks --> (+) whoziewhats, exemintes --> (+) maxabizers, frimbulators --> (-) whatajigs, funkados --> (+) frimbulators, funkados --> (-) maxabizers, marticatenes --> (+) refluppers, maxabizers --> (+) funkados, maxabizers --> (+) marticatenes, refluppers --> (+) exemintes, whatajigs --> (+) balacks, whoziewhats --> (+) funkados: Expected 1 to be 0.

  Message:
    Incorrect polarity discovered: Expected '-' to be '+'.

  Message:
    Incorrect polarity discovered: Expected '-' to be '+'.

  Message:
    Incorrect polarity discovered: Expected '+' to be '-'.

  Message:
    Incorrect polarity discovered: Expected '-' to be '+'.

  Message:
    Incorrect polarity discovered: Expected '-' to be '+'.

30) gemini-1.5-flash | causal translation testing | multiple feedback loops | extract 3 feedback loops with [+, +, -] polarities
  Message:
    Incorrect polarity discovered: Expected '-' to be '+'.

  Message:
    Incorrect polarity discovered: Expected '-' to be '+'.

Message:
    Incorrect polarity discovered: Expected '-' to be '+'.

  Message:
    Incorrect polarity discovered: Expected '-' to be '+'.

  Message:
    Incorrect polarity discovered: Expected '-' to be '+'.

31) gpt-4o | causal translation testing | single feedback loop | extract a reinforcing feedback loop with 3 variables
  Message:
    Incorrect polarity discovered: Expected '-' to be '+'.

  Message:
    Incorrect polarity discovered: Expected '-' to be '+'.

32) gpt-4o | causal translation testing | single feedback loop | extract a reinforcing feedback loop with 5 variables
  Message:
    Incorrect polarity discovered: Expected '-' to be '+'.

  Message:
    Incorrect polarity discovered: Expected '-' to be '+'.

  Message:
    Incorrect polarity discovered: Expected '-' to be '+'.

33) gpt-4o | causal translation testing | multiple feedback loops | extract 3 feedback loops with [-, -, +] polarities
  Message:
    Incorrect polarity discovered: Expected '-' to be '+'.

  Message:
    Incorrect polarity discovered: Expected '-' to be '+'.

34) gpt-4o | causal translation testing | single feedback loop | extract a balancing feedback loop with 7 variables
  Message:
    Incorrect polarity discovered: Expected '-' to be '+'.

  Message:
    Incorrect polarity discovered: Expected '-' to be '+'.

  Message:

Incorrect polarity discovered: Expected '-' to be '+'.

35) gpt-4o | causal translation testing | multiple feedback loops | extract 5 feedback loops with [-, +, +, +, -] polarities
  Message:
    Incorrect polarity discovered: Expected '-' to be '+'.

  Message:
    Incorrect polarity discovered: Expected '-' to be '+'.

  Message:
    Incorrect polarity discovered: Expected '-' to be '+'.

  Message:
    Incorrect polarity discovered: Expected '-' to be '+'.

  Message:
    Incorrect polarity discovered: Expected '-' to be '+'.

  Message:
    Incorrect polarity discovered: Expected '-' to be '+'.

  Message:
    Incorrect polarity discovered: Expected '-' to be '+'.

  Message:
    Incorrect polarity discovered: Expected '-' to be '+'.

36) gpt-4o | causal translation testing | multiple feedback loops | extract 2 feedback loops with [-, +] polarities
  Message:
    Incorrect polarity discovered: Expected '-' to be '+'.

  Message:
    Incorrect polarity discovered: Expected '-' to be '+'.

  Message:
    Incorrect polarity discovered: Expected '-' to be '+'.

  Message:
    Incorrect polarity discovered: Expected '-' to be '+'.

37) gpt-4o | causal translation testing | single feedback loop | extract a reinforcing feedback loop with 6 variables
  Message:
    Incorrect polarity discovered: Expected '-' to be '+'.

Message:
   Incorrect polarity discovered: Expected '-' to be '+'.

Message:
   Incorrect polarity discovered: Expected '-' to be '+'.

38) gpt-4o | causal translation testing | single feedback loop | extract a balancing feedback loop with 5 variables
   Message:
      Incorrect polarity discovered: Expected '-' to be '+'.

   Message:
      Incorrect polarity discovered: Expected '-' to be '+'.

39) gpt-4o | causal translation testing | single feedback loop | extract a balancing feedback loop with 8 variables
   Message:
      Incorrect polarity discovered: Expected '-' to be '+'.

   Message:
      Incorrect polarity discovered: Expected '-' to be '+'.

   Message:
      Incorrect polarity discovered: Expected '-' to be '+'.

40) gpt-4o | causal translation testing | single feedback loop | extract a reinforcing feedback loop with 7 variables
   Message:
      Incorrect polarity discovered: Expected '-' to be '+'.

   Message:
      Incorrect polarity discovered: Expected '-' to be '+'.

   Message:
      Incorrect polarity discovered: Expected '-' to be '+'.

   Message:
      Incorrect polarity discovered: Expected '-' to be '+'.

41) gpt-4o | causal translation testing | single feedback loop | extract a balancing feedback loop with 3 variables
   Message:
      Incorrect polarity discovered: Expected '-' to be '+'.

42) gpt-4o | causal translation testing | single feedback loop | extract a balancing feedback loop with 6 variables
  Message:
    Incorrect polarity discovered: Expected '-' to be '+'.

  Message:
    Incorrect polarity discovered: Expected '-' to be '+'.

43) gpt-4o | causal translation testing | single feedback loop | extract a reinforcing feedback loop with 8 variables
  Message:
    Incorrect polarity discovered: Expected '-' to be '+'.

  Message:
    Incorrect polarity discovered: Expected '-' to be '+'.

  Message:
    Incorrect polarity discovered: Expected '-' to be '+'.

  Message:
    Incorrect polarity discovered: Expected '-' to be '+'.

44) gpt-4o | causal translation testing | multiple feedback loops | extract 5 feedback loops with [-, +, +, -, -] polarities
  Message:
    Incorrect polarity discovered: Expected '-' to be '+'.

  Message:
    Incorrect polarity discovered: Expected '-' to be '+'.

  Message:
    Incorrect polarity discovered: Expected '-' to be '+'.

  Message:
    Incorrect polarity discovered: Expected '-' to be '+'.

  Message:
    Incorrect polarity discovered: Expected '-' to be '+'.

  Message:
    Incorrect polarity discovered: Expected '-' to be '+'.

  Message:
    Incorrect polarity discovered: Expected '-' to be '+'.

45) gpt-4o-mini | causal translation testing | single feedback loop | extract a balancing feedback loop with 3 variables
  Message:
    Incorrect polarity discovered: Expected '-' to be '+'.

  Message:
    Incorrect polarity discovered: Expected '+' to be '-'.

46) gpt-4o-mini | causal translation testing | multiple feedback loops | extract 5 feedback loops with [-, +, +, -, -] polarities
  Message:
    Incorrect polarity discovered: Expected '-' to be '+'.

  Message:
    Incorrect polarity discovered: Expected '-' to be '+'.

  Message:
    Incorrect polarity discovered: Expected '-' to be '+'.

  Message:
    Incorrect polarity discovered: Expected '-' to be '+'.

  Message:
    Incorrect polarity discovered: Expected '+' to be '-'.

  Message:
    Incorrect polarity discovered: Expected '-' to be '+'.

  Message:
    Incorrect polarity discovered: Expected '-' to be '+'.

  Message:
    Incorrect polarity discovered: Expected '-' to be '+'.

  Message:
    Incorrect polarity discovered: Expected '-' to be '+'.

  Message:
    Incorrect polarity discovered: Expected '+' to be '-'.

  Message:
    Incorrect polarity discovered: Expected '-' to be '+'.

47) gpt-4o-mini | causal translation testing | single feedback loop | extract a balancing feedback loop with 6 variables
  Message:
    Fake relationships found
    Marticatenes --> (+) Priaries
    Ground Truth
    exemintes --> (+) ocs, marticatenes --> (-) refluppers, ocs --> (+) proptimatires, priaries --> (+) marticatenes, proptimatires --> (+) priaries, refluppers --> (+) exemintes: Expected 1 to be 0.

  Message:
    Incorrect polarity discovered: Expected '-' to be '+'.

  Message:
    Incorrect polarity discovered: Expected '+' to be '-'.

  Message:
    Incorrect polarity discovered: Expected '-' to be '+'.

48) gpt-4o-mini | causal translation testing | multiple feedback loops | extract 5 feedback loops with [-, +, +, +, -] polarities
  Message:
    Incorrect polarity discovered: Expected '-' to be '+'.

  Message:
    Incorrect polarity discovered: Expected '-' to be '+'.

  Message:
    Incorrect polarity discovered: Expected '-' to be '+'.

  Message:
    Incorrect polarity discovered: Expected '-' to be '+'.

  Message:
    Incorrect polarity discovered: Expected '+' to be '-'.

  Message:
    Incorrect polarity discovered: Expected '-' to be '+'.

  Message:
    Incorrect polarity discovered: Expected '+' to be '-'.

  Message:
    Incorrect polarity discovered: Expected '-' to be '+'.

  Message:
    Incorrect polarity discovered: Expected '-' to be '+'.

Message:
    Incorrect polarity discovered: Expected '-' to be '+'.

  Message:
    Incorrect polarity discovered: Expected '-' to be '+'.

49) gpt-4o-mini | causal translation testing | single feedback loop | extract a balancing feedback loop with 4 variables
  Message:
    Fake relationships found
    funkados --> (-) refluppers, maxabizers --> (-) funkados
    Ground Truth
    funkados --> (-) maxabizers, marticatenes --> (+) refluppers, maxabizers --> (+) marticatenes, refluppers --> (+) funkados: Expected 2 to be 0.

  Message:
    Incorrect polarity discovered: Expected '+' to be '-'.

  Message:
    Incorrect polarity discovered: Expected '-' to be '+'.

50) gpt-4o-mini | causal translation testing | multiple feedback loops | extract 3 feedback loops with [-, -, +] polarities
  Message:
    Incorrect polarity discovered: Expected '-' to be '+'.

  Message:
    Incorrect polarity discovered: Expected '+' to be '-'.

  Message:
    Incorrect polarity discovered: Expected '-' to be '+'.

  Message:
    Incorrect polarity discovered: Expected '+' to be '-'.

  Message:
    Incorrect polarity discovered: Expected '-' to be '+'.

  Message:
    Incorrect polarity discovered: Expected '-' to be '+'.

51) gpt-4o-mini | causal translation testing | multiple feedback loops | extract 3 feedback loops with [+, +, -] polarities
  Message:
    Incorrect polarity discovered: Expected '-' to be '+'.

Message:
     Incorrect polarity discovered: Expected '-' to be '+'.

   Message:
     Incorrect polarity discovered: Expected '-' to be '+'.

52) gpt-4o-mini | causal translation testing | single feedback loop | extract a reinforcing feedback loop with 5 variables
   Message:
     Fake relationships found
     marticatenes --> (+) maxabizers
     Ground Truth
     exemintes --> (+) ocs, marticatenes --> (+) refluppers, maxabizers --> (+) marticatenes, ocs --> (+) maxabizers, refluppers --> (+) exemintes: Expected 1 to be 0.

   Message:
     Incorrect polarity discovered: Expected '-' to be '+'.

   Message:
     Incorrect polarity discovered: Expected '-' to be '+'.

   Message:
     Incorrect polarity discovered: Expected '-' to be '+'.

53) gpt-4o-mini | causal translation testing | single feedback loop | extract a reinforcing feedback loop with 2 variables
   Message:
     Incorrect polarity discovered: Expected '-' to be '+'.

54) gpt-4o-mini | causal translation testing | single feedback loop | extract a balancing feedback loop with 5 variables
   Message:
     Fake relationships found
     Maxabizers --> (-) Ocs
     Ground Truth
     exemintes --> (+) ocs, marticatenes --> (+) refluppers, maxabizers --> (-) marticatenes, ocs --> (+) maxabizers, refluppers --> (+) exemintes: Expected 1 to be 0.

   Message:
     Incorrect polarity discovered: Expected '-' to be '+'.

   Message:
     Incorrect polarity discovered: Expected '-' to be '+'.

55) gpt-4o-mini | causal translation testing | multiple feedback loops | extract 2 feedback loops with [-, +] polarities
  Message:
    Incorrect polarity discovered: Expected '-' to be '+'.

  Message:
    Incorrect polarity discovered: Expected '-' to be '+'.

  Message:
    Incorrect polarity discovered: Expected '+' to be '-'.

  Message:
    Incorrect polarity discovered: Expected '-' to be '+'.

  Message:
    Incorrect polarity discovered: Expected '-' to be '+'.

56) gpt-4o-mini | causal translation testing | single feedback loop | extract a reinforcing feedback loop with 7 variables
  Message:
    Fake relationships found
    povals --> (+) exemintes, refluppers --> (+) povals
    Ground Truth
    exemintes --> (+) ocs, houtals --> (+) povals, ocs --> (+) proptimatires, povals --> (+) refluppers, priaries --> (+) houtals, proptimatires --> (+) priaries, refluppers --> (+) exemintes: Expected 2 to be 0.

  Message:
    Incorrect polarity discovered: Expected '-' to be '+'.

  Message:
    Incorrect polarity discovered: Expected '-' to be '+'.

  Message:
    Incorrect polarity discovered: Expected '-' to be '+'.

57) gpt-4o-mini | causal translation testing | multiple feedback loops | extract 2 feedback loops with [+, +] polarities
  Message:
    Incorrect polarity discovered: Expected '-' to be '+'.

  Message:
    Incorrect polarity discovered: Expected '-' to be '+'.

  Message:
    Incorrect polarity discovered: Expected '-' to be '+'.

  Message:
    Incorrect polarity discovered: Expected '-' to be '+'.

58) gpt-4o-mini | causal translation testing | single feedback loop | extract a reinforcing feedback loop with 6 variables
  Message:
    Fake relationships found
    Marticatenes --> (+) Priaries
    Ground Truth
    exemintes --> (+) ocs, marticatenes --> (+) refluppers, ocs --> (+) proptimatires, priaries --> (+) marticatenes, proptimatires --> (+) priaries, refluppers --> (+) exemintes: Expected 1 to be 0.

  Message:
    Incorrect polarity discovered: Expected '-' to be '+'.

  Message:
    Incorrect polarity discovered: Expected '-' to be '+'.

59) gpt-4o-mini | causal translation testing | single feedback loop | extract a balancing feedback loop with 8 variables
  Message:
    Fake relationships found
    dominitoxings --> (+) ocs, exemintes --> (+) dominitoxings
    Ground Truth
    auspongs --> (+) dominitoxings, dominitoxings --> (+) exemintes, exemintes --> (-) ocs, houtals --> (+) povals, ocs --> (+) proptimatires, povals --> (+) auspongs, priaries --> (+) houtals, proptimatires --> (+) priaries: Expected 2 to be 0.

  Message:
    Incorrect polarity discovered: Expected '-' to be '+'.

  Message:
    Incorrect polarity discovered: Expected '+' to be '-'.

  Message:
    Incorrect polarity discovered: Expected '-' to be '+'.

  Message:
    Incorrect polarity discovered: Expected '-' to be '+'.

60) gpt-4o-mini | causal translation testing | single feedback loop | extract a reinforcing feedback loop with 3 variables
  Message:
    Fake relationships found

maxabizers --> (-) funkados
    Ground Truth
    funkados --> (+) maxabizers, maxabizers --> (+) whoziewhats, whoziewhats --> (+) funkados: Expected 1 to be 0.

  Message:
    Incorrect polarity discovered: Expected '-' to be '+'.

  Message:
    Incorrect polarity discovered: Expected '-' to be '+'.

61) gpt-4o-mini | causal translation testing | single feedback loop | extract a reinforcing feedback loop with 8 variables
  Message:
    Fake relationships found
    exemintes --> (+) dominitoxings
    Ground Truth
    auspongs --> (+) dominitoxings, dominitoxings --> (+) exemintes, exemintes --> (+) ocs, houtals --> (+) povals, ocs --> (+) proptimatires, povals --> (+) auspongs, priaries --> (+) houtals, proptimatires --> (+) priaries: Expected 1 to be 0.

  Message:
    Incorrect polarity discovered: Expected '-' to be '+'.

  Message:
    Incorrect polarity discovered: Expected '-' to be '+'.

  Message:
    Incorrect polarity discovered: Expected '-' to be '+'.

62) gpt-4o-mini | causal translation testing | single feedback loop | extract a balancing feedback loop with 7 variables
  Message:
    Fake relationships found
    refluppers --> (+) povals
    Ground Truth
    exemintes --> (+) ocs, houtals --> (+) povals, ocs --> (+) proptimatires, povals --> (+) refluppers, priaries --> (+) houtals, proptimatires --> (+) priaries, refluppers --> (-) exemintes: Expected 1 to be 0.

  Message:
    Incorrect polarity discovered: Expected '-' to be '+'.

  Message:
    Incorrect polarity discovered: Expected '-' to be '+'.

  Message:

Incorrect polarity discovered: Expected '-' to be '+'.

  Message:
    Incorrect polarity discovered: Expected '+' to be '-'.

63) gpt-4o-mini | causal translation testing | single relationship | extract a reinforcing relationship up
  Message:
    Fake relationships found
    Whatajigs --> (+) Frimbulators
    Ground Truth
    frimbulators --> (+) whatajigs: Expected 1 to be 0.

64) gpt-4o-mini | causal translation testing | single relationship | extract a balancing relationship down
  Message:
    Fake relationships found
    Whatajigs --> (-) Frimbulators
    Ground Truth
    frimbulators --> (-) whatajigs: Expected 1 to be 0.

  Message:
    Incorrect polarity discovered: Expected '+' to be '-'.

65) gpt-4o-mini | causal translation testing | single feedback loop | extract a reinforcing feedback loop with 4 variables
  Message:
    Fake relationships found
    marticatenes --> (-) maxabizers
    Ground Truth
    funkados --> (+) maxabizers, marticatenes --> (+) refluppers, maxabizers --> (+) marticatenes, refluppers --> (+) funkados: Expected 1 to be 0.

  Message:
    Incorrect polarity discovered: Expected '-' to be '+'.

  Message:
    Incorrect polarity discovered: Expected '-' to be '+'.

66) gpt-4o-mini | causal translation testing | single relationship | extract a reinforcing relationship down
  Message:
    Fake relationships found
    Whatajigs --> (+) Frimbulators
    Ground Truth
    frimbulators --> (+) whatajigs: Expected 1 to be 0.

Message:
  Incorrect polarity discovered: Expected '-' to be '+'.

67) gpt-4o-mini | causal translation testing | single relationship | extract a balancing relationship up
  Message:
    Fake relationships found
    Frimbulators --> (-) Whatajig Demand, Whatajig Demand --> (+) Frimbulators, Whatajigs --> (+) Frimbulators
    Ground Truth
    frimbulators --> (-) whatajigs: Expected 3 to be 0.

68) o3-mini high | causal translation testing | single relationship | extract a balancing relationship down
  Message:
    Fake relationships found
    Whatajigs --> (-) Frimbulators
    Ground Truth
    frimbulators --> (-) whatajigs: Expected 1 to be 0.

69) o3-mini high | causal translation testing | single relationship | extract a reinforcing relationship down
  Message:
    Fake relationships found
    Whatajigs --> (+) Frimbulators
    Ground Truth
    frimbulators --> (+) whatajigs: Expected 1 to be 0.

70) o3-mini high | causal translation testing | single relationship | extract a balancing relationship up
  Message:
    Fake relationships found
    Whatajigs --> (-) Frimbulators
    Ground Truth
    frimbulators --> (-) whatajigs: Expected 1 to be 0.

71) o3-mini medium | causal translation testing | single relationship | extract a balancing relationship up
  Message:
    Fake relationships found
    Frimbulator Count --> (-) Whatajig Count, Whatajig Count --> (-) Frimbulator Count
    Ground Truth
    frimbulators --> (-) whatajigs: Expected 2 to be 0.

  Message:
    Real relationships not found
    frimbulators --> (-) whatajigs
    Ground Truth
    frimbulators --> (-) whatajigs: Expected 1 to be 0.

72) o3-mini medium | causal translation testing | single relationship | extract a balancing relationship down
   Message:
     Fake relationships found
     Whatajigs --> (-) Frimbulators
     Ground Truth
     frimbulators --> (-) whatajigs: Expected 1 to be 0.

73) gemini-2.0-flash | causal translation testing | single feedback loop | extract a balancing feedback loop with 4 variables
   Message:
     Fake relationships found
     marticatenes --> (-) maxabizers
     Ground Truth
     funkados --> (-) maxabizers, marticatenes --> (+) refluppers, maxabizers --> (+) marticatenes, refluppers --> (+) funkados: Expected 1 to be 0.

74) gemini-2.0-flash | causal translation testing | single relationship | extract a reinforcing relationship up
   Message:
     Fake relationships found
     whatajigs --> (+) frimbulators
     Ground Truth
     frimbulators --> (+) whatajigs: Expected 1 to be 0.

75) gemini-2.0-flash | causal translation testing | single feedback loop | extract a reinforcing feedback loop with 8 variables
   Message:
     Incorrect polarity discovered: Expected '-' to be '+'.

   Message:
     Incorrect polarity discovered: Expected '-' to be '+'.

   Message:
     Incorrect polarity discovered: Expected '-' to be '+'.

   Message:
     Incorrect polarity discovered: Expected '-' to be '+'.

76) gemini-2.0-flash | causal translation testing | single relationship | extract a reinforcing relationship down
   Message:
     Fake relationships found
     Number of frimbulators --> (+) Number of whatajigs, Number of whatajigs --> (+) Number of frimbulators
     Ground Truth

   frimbulators --> (+) whatajigs: Expected 2 to be 0.

  Message:
   Real relationships not found
   frimbulators --> (+) whatajigs
   Ground Truth
   frimbulators --> (+) whatajigs: Expected 1 to be 0.

77) gemini-2.0-flash | causal translation testing | single feedback loop | extract a reinforcing feedback loop with 6 variables
  Message:
   Incorrect polarity discovered: Expected '-' to be '+'.

  Message:
   Incorrect polarity discovered: Expected '-' to be '+'.

  Message:
   Incorrect polarity discovered: Expected '-' to be '+'.

78) gemini-2.0-flash | causal translation testing | single feedback loop | extract a reinforcing feedback loop with 3 variables
  Message:
   Incorrect polarity discovered: Expected '-' to be '+'.

  Message:
   Incorrect polarity discovered: Expected '-' to be '+'.

79) gemini-2.0-flash | causal translation testing | single relationship | extract a balancing relationship up
  Message:
   Fake relationships found
   whatajigs --> (-) frimbulators
   Ground Truth
   frimbulators --> (-) whatajigs: Expected 1 to be 0.

80) gemini-2.0-flash | causal translation testing | single feedback loop | extract a reinforcing feedback loop with 4 variables
  Message:
   Incorrect polarity discovered: Expected '-' to be '+'.

  Message:
   Incorrect polarity discovered: Expected '-' to be '+'.

81) gemini-2.0-flash | causal translation testing | multiple feedback loops | extract 3 feedback loops with [+, +, -] polarities
  Message:
    Fake relationships found
    maxabizers --> (-) exemintes
    Ground Truth
    balacks --> (+) whoziewhats, exemintes --> (+) maxabizers, frimbulators --> (+) whatajigs, funkados --> (+) frimbulators, funkados --> (+) maxabizers, marticatenes --> (+) refluppers, maxabizers --> (+) funkados, maxabizers --> (-) marticatenes, refluppers --> (+) exemintes, whatajigs --> (+) balacks, whoziewhats --> (+) funkados: Expected 1 to be 0.

82) gemini-2.0-flash | causal translation testing | single relationship | extract a balancing relationship down
  Message:
    Fake relationships found
    Number of frimbulators --> (-) Number of whatajigs, Number of whatajigs --> (-) Number of frimbulators
    Ground Truth
    frimbulators --> (-) whatajigs: Expected 2 to be 0.

  Message:
    Real relationships not found
    frimbulators --> (-) whatajigs
    Ground Truth
    frimbulators --> (-) whatajigs: Expected 1 to be 0.

83) gemini-2.0-flash | causal translation testing | multiple feedback loops | extract 2 feedback loops with [+, +] polarities
  Message:
    Fake relationships found
    refluppers --> (-) marticatenes, whatajigs --> (-) frimbulators
    Ground Truth
    balacks --> (+) frimbulators, balacks --> (+) whoziewhats, frimbulators --> (+) whatajigs, funkados --> (+) maxabizers, marticatenes --> (+) refluppers, maxabizers --> (+) marticatenes, refluppers --> (+) balacks, whatajigs --> (+) balacks, whoziewhats --> (+) funkados: Expected 2 to be 0.

84) gemini-2.0-flash | causal translation testing | single feedback loop | extract a reinforcing feedback loop with 5 variables
  Message:
    Incorrect polarity discovered: Expected '-' to be '+'.

  Message:
    Incorrect polarity discovered: Expected '-' to be '+'.

  Message:
    Incorrect polarity discovered: Expected '-' to be '+'.

85) o1 | conformance testing | can conform to the instruction include a minimum number of feedback loops and a minimum number of variables| for the case American Revolution
    Message:
      Too few feedback loops: The number of feedback loops found was 1: Expected 1 to be greater than or equal 6.

86) o1 | conformance testing | can conform to the instruction include a minimum number of feedback loops| for the case American Revolution
    Message:
      Too few feedback loops: The number of feedback loops found was 6: Expected 6 to be greater than or equal 8.

87) o1 | conformance testing | can conform to the instruction include a maximum number of feedback loops| for the case Road Rage
    Message:
      Too many feedback loops: The number of feedback loops found was 9: Expected 9 to be less than or equal 4.

88) o1 | conformance testing | can conform to the instruction include a minimum number of feedback loops and a minimum number of variables| for the case Road Rage
    Message:
      Too few feedback loops: The number of feedback loops found was 5: Expected 5 to be greater than or equal 6.

89) o1 | conformance testing | can conform to the instruction include a min number of feedback loops and a maximum number of variables| for the case American Revolution
    Message:
      Too few feedback loops: The number of feedback loops found was 3: Expected 3 to be greater than or equal 6.

90) o3-mini medium | conformance testing | can conform to the instruction include a min number of feedback loops and a maximum number of variables| for the case Road Rage
    Message:
      Too few feedback loops: The number of feedback loops found was 4: Expected 4 to be greater than or equal 6.

91) o3-mini medium | conformance testing | can conform to the instruction include a maximum number of feedback loops and a minimum number of variables| for the case Road Rage
    Message:
      Too many feedback loops: The number of feedback loops found was 5: Expected 5 to be less than or equal 4.

92) o3-mini medium | conformance testing | can conform to the instruction include a minimum number of feedback loops| for the case Road Rage
   Message:
    Too few feedback loops: The number of feedback loops found was 5: Expected 5 to be greater than or equal 8.

93) o3-mini medium | conformance testing | can conform to the instruction include a min number of feedback loops and a maximum number of variables| for the case American Revolution
   Message:
    Too few feedback loops: The number of feedback loops found was 4: Expected 4 to be greater than or equal 6.

94) o3-mini medium | conformance testing | can conform to the instruction include a minimum number of feedback loops and a minimum number of variables| for the case American Revolution
   Message:
    Too few feedback loops: The number of feedback loops found was 3: Expected 3 to be greater than or equal 6.

95) o3-mini medium | conformance testing | can conform to the instruction include a minimum number of feedback loops| for the case American Revolution
   Message:
    Too few feedback loops: The number of feedback loops found was 3: Expected 3 to be greater than or equal 8.

96) o3-mini high | conformance testing | can conform to the instruction include a min number of feedback loops and a maximum number of variables| for the case Road Rage
   Message:
    Too few feedback loops: The number of feedback loops found was 5: Expected 5 to be greater than or equal 6.

97) o3-mini high | conformance testing | can conform to the instruction include a minimum number of feedback loops| for the case Road Rage
   Message:
    Too few feedback loops: The number of feedback loops found was 6: Expected 6 to be greater than or equal 8.

98) o3-mini high | conformance testing | can conform to the instruction include a min number of feedback loops and a maximum number of variables| for the case American Revolution
   Message:

    Too few feedback loops: The number of feedback loops found was 4: Expected 4 to be greater than or equal 6.

99) o3-mini high | conformance testing | can conform to the instruction include a minimum number of feedback loops and a minimum number of variables| for the case Road Rage
   Message:
    Too few feedback loops: The number of feedback loops found was 5: Expected 5 to be greater than or equal 6.

100) o3-mini high | conformance testing | can conform to the instruction include a minimum number of feedback loops and a minimum number of variables| for the case American Revolution
   Message:
    Too few feedback loops: The number of feedback loops found was 5: Expected 5 to be greater than or equal 6.

101) o3-mini high | conformance testing | can conform to the instruction include a minimum number of feedback loops| for the case American Revolution
   Message:
    Too few feedback loops: The number of feedback loops found was 3: Expected 3 to be greater than or equal 8.

102) gemini-1.5-flash | conformance testing | can conform to the instruction include a maximum number of variables| for the case Road Rage
   Message:
    Too many variables: Variables are: Stress Levels, Aggressive Driving Incidents, Traffic Congestion, Road Design Improvements, Stress Reduction Techniques, Public Awareness Campaigns: Expected 6 to be less than or equal 5.

103) gemini-1.5-flash | conformance testing | can conform to the instruction include a minimum number of feedback loops| for the case Road Rage
   Message:
    Too few feedback loops: The number of feedback loops found was 0: Expected 0 to be greater than or equal 8.

104) gemini-1.5-flash | conformance testing | can conform to the instruction include a min number of feedback loops and a maximum number of variables| for the case Road Rage
   Message:
    Too few feedback loops: The number of feedback loops found was 4: Expected 4 to be greater than or equal 6.

105) gemini-1.5-flash | conformance testing | can conform to the instruction include a minimum number of feedback loops| for the case American Revolution
   Message:
    Too few feedback loops: The number of feedback loops found was 3: Expected 3 to be greater than or equal 8.

106) gemini-1.5-flash | conformance testing | can conform to the instruction include a minimum number of feedback loops and a minimum number of variables| for the case American Revolution
   Message:
    Too few feedback loops: The number of feedback loops found was 5: Expected 5 to be greater than or equal 6.

107) gemini-1.5-flash | conformance testing | can conform to the instruction include a min number of feedback loops and a maximum number of variables| for the case American Revolution
   Message:
    Too few feedback loops: The number of feedback loops found was 0: Expected 0 to be greater than or equal 6.

108) gemini-1.5-flash | conformance testing | can conform to the instruction include a maximum number of variables| for the case American Revolution
   Message:
    Too many variables: Variables are: British Policies (Taxation, Repression), Colonial Resistance, British Repressive Measures, Colonial Unity, American Revolution, Colonial Identity, French and Indian War: Expected 7 to be less than or equal 5.

109) gemini-2.0-flash-lite | conformance testing | can conform to the instruction include a minimum number of feedback loops and a minimum number of variables| for the case Road Rage
   Message:
    Too few variables: Variables are: Stress Levels, Road Rage Incidents, Traffic Congestion, Poor Road Design, Perceived Provocations, Learned Behavior, Public Awareness of Road Rage: Expected 7 to be greater than or equal 8.

   Message:
    Too few feedback loops: The number of feedback loops found was 5: Expected 5 to be greater than or equal 6.

110) gemini-2.0-flash-lite | conformance testing | can conform to the instruction include a maximum number of feedback loops| for the case Road Rage
   Message:
    Too many feedback loops: The number of feedback loops found was 9: Expected 9 to be less than or equal 4.

111) gemini-2.0-flash-lite | conformance testing | can conform to the instruction include a minimum number of feedback loops| for the case American Revolution
  Message:
    Too few feedback loops: The number of feedback loops found was 5: Expected 5 to be greater than or equal 8.

112) gemini-2.0-flash-lite | conformance testing | can conform to the instruction include a maximum number of feedback loops and a maximum number of variables| for the case American Revolution
  Message:
    Too many variables: Variables are: British Taxation, Colonial Resentment, Colonial Defiance, Anti-British Sentiment, British Response, French and Indian War, Colonial Identity: Expected 7 to be less than or equal 5.

113) gemini-2.0-flash-lite | conformance testing | can conform to the instruction include a minimum number of variables| for the case American Revolution
  Message:
    Too few variables: Variables are: British Taxation, Colonial Economic Hardship, Colonial Satisfaction with British Rule, Colonial Defiance, British Response to Colonial Defiance, Colonial Anti-British Sentiment, Colonial Identity, French and Indian War Debt, British Military Presence: Expected 9 to be greater than or equal 10.

114) gemini-2.0-flash-lite | conformance testing | can conform to the instruction include a min number of feedback loops and a maximum number of variables| for the case American Revolution
  Message:
    Too few feedback loops: The number of feedback loops found was 5: Expected 5 to be greater than or equal 6.

115) gemini-2.0-flash-lite | conformance testing | can conform to the instruction include a minimum number of feedback loops and a minimum number of variables| for the case American Revolution
  Message:
    Too few variables: Variables are: British Taxation, Colonial Grievances, Colonial Identity, Acts of Defiance, British Response, Revolutionary Sentiment, Armed Conflict: Expected 7 to be greater than or equal 8.

  Message:
    Too few feedback loops: The number of feedback loops found was 3: Expected 3 to be greater than or equal 6.

116) gpt-4o-mini | conformance testing | can conform to the instruction include a min number of feedback loops and a maximum number of variables| for the case American Revolution
  Message:
    Too few feedback loops: The number of feedback loops found was 5: Expected 5 to be greater than or equal 6.

117) gpt-4o-mini | conformance testing | can conform to the instruction include a minimum number of variables| for the case Road Rage
   Message:
    Too few variables: Variables are: High Stress Levels, Impatience, Aggressive Driving, Perceived Provocations, Traffic Congestion, Weather Conditions, Learned Behavior, Lack of Sleep, Distracted Driving: Expected 9 to be greater than or equal 10.

118) gpt-4o-mini | conformance testing | can conform to the instruction include a minimum number of feedback loops| for the case Road Rage
   Message:
    Too few feedback loops: The number of feedback loops found was 7: Expected 7 to be greater than or equal 8.

119) gpt-4o-mini | conformance testing | can conform to the instruction include a minimum number of variables| for the case American Revolution
   Message:
    Too few variables: Variables are: Taxation, Colonial Anger, Anti-British Sentiment, Boston Massacre, Acts of Defiance, Colonial Identity, Desire for Independence, Intolerable Acts: Expected 8 to be greater than or equal 10.

120) gpt-4o-mini | conformance testing | can conform to the instruction include a minimum number of feedback loops| for the case American Revolution
   Message:
    Too few feedback loops: The number of feedback loops found was 5: Expected 5 to be greater than or equal 8.

121) gemini-2.0-flash | conformance testing | can conform to the instruction include a min number of feedback loops and a maximum number of variables| for the case American Revolution
   Message:
    Too few feedback loops: The number of feedback loops found was 4: Expected 4 to be greater than or equal 6.

122) gemini-2.0-flash | conformance testing | can conform to the instruction include a maximum number of feedback loops| for the case Road Rage
   Message:
    Too many feedback loops: The number of feedback loops found was 5: Expected 5 to be less than or equal 4.

123) gemini-2.0-flash | conformance testing | can conform to the instruction include a maximum number of variables| for the case Road Rage
   Message:

    Too many variables: Variables are: Traffic Congestion, Driver Frustration, Road Rage Incidents, Technology Use While Driving, Distracted Driving, Perceived Provocations, Stress and Anxiety, Law Enforcement: Expected 8 to be less than or equal 5.

124) gemini-2.0-flash | conformance testing | can conform to the instruction include a maximum number of feedback loops and a maximum number of variables| for the case Road Rage
    Message:
    Too many variables: Variables are: Traffic Congestion, Driver Stress, Aggressive Driving, Perceived Safety, Cautious Driving, Societal Stress, Road Rage Incidents, Law Enforcement: Expected 8 to be less than or equal 5.

    Message:
    Too many feedback loops: The number of feedback loops found was 5: Expected 5 to be less than or equal 4.

125) gemini-2.0-flash | conformance testing | can conform to the instruction include a minimum number of feedback loops and a minimum number of variables| for the case American Revolution
    Message:
    Too few feedback loops: The number of feedback loops found was 4: Expected 4 to be greater than or equal 6.

126) gemini-2.0-flash | conformance testing | can conform to the instruction include a maximum number of variables| for the case American Revolution
    Message:
    Too many variables: Variables are: Taxation, Colonial Discontent, Boston Massacre, Anti-British Sentiment, Boston Tea Party, Tensions with Britain, Intolerable Acts, Colonial Resistance, Colonial Identity: Expected 9 to be less than or equal 5.

127) gemini-2.0-flash | conformance testing | can conform to the instruction include a maximum number of feedback loops and a maximum number of variables| for the case American Revolution
    Message:
    Too many variables: Variables are: Taxation, Colonial Resistance, British Revenue, British Actions, Colonial Anger, Colonial Identity, Boston Massacre, Anti-British Sentiment: Expected 8 to be less than or equal 5.

128) gemini-2.0-flash | conformance testing | can conform to the instruction include a minimum number of feedback loops| for the case American Revolution
    Message:
    Too few feedback loops: The number of feedback loops found was 5: Expected 5 to be greater than or equal 8.

129) gpt-4o | conformance testing | can conform to the instruction include a min number of feedback loops and a maximum number of variables| for the case American Revolution

130) gpt-4o | conformance testing | can conform to the instruction include a maximum number of feedback loops and a minimum number of variables| for the case Road Rage
   Message:
    Too many feedback loops: The number of feedback loops found was 5: Expected 5 to be less than or equal 4.

131) gpt-4o | conformance testing | can conform to the instruction include a minimum number of feedback loops| for the case American Revolution
   Message:
    Too few feedback loops: The number of feedback loops found was 3: Expected 3 to be greater than or equal 8.

132) gpt-4o | conformance testing | can conform to the instruction include a minimum number of feedback loops and a minimum number of variables| for the case American Revolution
   Message:
    Too few feedback loops: The number of feedback loops found was 2: Expected 2 to be greater than or equal 6.

133) gpt-4.5-preview | conformance testing | can conform to the instruction include a minimum number of variables| for the case American Revolution
   Message:
    Too few variables: Variables are: British taxation, Colonial anger, Colonial resistance, British punitive measures, Boston Massacre, Boston Tea Party, Intolerable Acts, French and Indian War debt, Colonial identity: Expected 9 to be greater than or equal 10.

134) gpt-4.5-preview | conformance testing | can conform to the instruction include a min number of feedback loops and a maximum number of variables| for the case American Revolution
   Message:
    Too few feedback loops: The number of feedback loops found was 5: Expected 5 to be greater than or equal 6.

135) gpt-4.5-preview | conformance testing | can conform to the instruction include a minimum number of feedback loops| for the case American Revolution
   Message:
    Too few feedback loops: The number of feedback loops found was 3: Expected 3 to be greater than or equal 8.

135) llama3.3:70b-instruct-q5_K_M | causal translation testing | single relationship | extract a balancing relationship down
  Message:
    Fake relationships found
    Whatajigs --> (-) Frimbulators
    Ground Truth
    frimbulators --> (-) whatajigs: Expected 1 to be 0.

136) llama3.3:70b-instruct-q5_K_M | causal translation testing | single feedback loop | extract a balancing feedback loop with 5 variables
  Message:
    Fake relationships found
    exemintes --> (-) maxabizers, marticatenes --> (+) exemintes, marticatenes --> (+) ocs, maxabizers --> (-) exemintes, maxabizers --> (-) refluppers, ocs --> (-) marticatenes, ocs --> (-) refluppers, refluppers --> (-) maxabizers
    Ground Truth
    exemintes --> (+) ocs, marticatenes --> (+) refluppers, maxabizers --> (-) marticatenes, ocs --> (+) maxabizers, refluppers --> (+) exemintes: Expected 8 to be 0.

  Message:
    Incorrect polarity discovered: Expected '-' to be '+'.

137) llama3.3:70b-instruct-q5_K_M | causal translation testing | single feedback loop | extract a balancing feedback loop with 4 variables
  Message:
    Incorrect polarity discovered: Expected '-' to be '+'.

138) llama3.3:70b-instruct-q5_K_M | causal translation testing | single feedback loop | extract a balancing feedback loop with 8 variables
  Message:
    Fake relationships found
    auspongs --> (-) exemintes, auspongs --> (-) ocs, dominitoxings --> (-) ocs, exemintes --> (+) dominitoxings, houtals --> (-) exemintes, houtals --> (-) ocs, ocs --> (-) exemintes, povals --> (-) exemintes, povals --> (-) ocs, priaries --> (-) exemintes, priaries --> (-) ocs, proptimatires --> (-) exemintes, proptimatires --> (-) ocs
    Ground Truth
    auspongs --> (+) dominitoxings, dominitoxings --> (+) exemintes, exemintes --> (-) ocs, houtals --> (+) povals, ocs --> (+) proptimatires, povals --> (+) auspongs, priaries --> (+) houtals, proptimatires --> (+) priaries: Expected 13 to be 0.

139) llama3.3:70b-instruct-q5_K_M | causal translation testing | multiple feedback loops | extract 2 feedback loops with [-, +] polarities
  Message:
    Fake relationships found

    B --> (+) F, B --> (+) H, D --> (+) H, D --> (+) M, F --> (+) B, F --> (-) W, H --> (+) D, M --> (+) T, R --> (+) B, R --> (-) T, T --> (-) M, T --> (-) R, W --> (-) F
    Ground Truth
    balacks --> (+) frimbulators, balacks --> (+) whoziewhats, frimbulators --> (-) whatajigs, funkados --> (+) maxabizers, marticatenes --> (+) refluppers, maxabizers --> (+) marticatenes, refluppers --> (+) balacks, whatajigs --> (+) balacks, whoziewhats --> (+) funkados: Expected 13 to be 0.

  Message:
    Real relationships not found
    balacks --> (+) frimbulators, balacks --> (+) whoziewhats, frimbulators --> (-) whatajigs, funkados --> (+) maxabizers, marticatenes --> (+) refluppers, maxabizers --> (+) marticatenes, refluppers --> (+) balacks, whatajigs --> (+) balacks, whoziewhats --> (+) funkados
    Ground Truth
    balacks --> (+) frimbulators, balacks --> (+) whoziewhats, frimbulators --> (-) whatajigs, funkados --> (+) maxabizers, marticatenes --> (+) refluppers, maxabizers --> (+) marticatenes, refluppers --> (+) balacks, whatajigs --> (+) balacks, whoziewhats --> (+) funkados: Expected 9 to be 0.

140) llama3.3:70b-instruct-q5_K_M | causal translation testing | single feedback loop | extract a reinforcing feedback loop with 7 variables
  Message:
    Fake relationships found
    exemintes --> (+) houtals, exemintes --> (+) proptimatires, exemintes --> (-) povals, exemintes --> (-) priaries, priaries --> (+) povals, proptimatires --> (+) houtals, proptimatires --> (-) povals
    Ground Truth
    exemintes --> (+) ocs, houtals --> (+) povals, ocs --> (+) proptimatires, povals --> (+) refluppers, priaries --> (+) houtals, proptimatires --> (+) priaries, refluppers --> (+) exemintes: Expected 7 to be 0.

  Message:
    Incorrect polarity discovered: Expected '-' to be '+'.

  Message:
    Incorrect polarity discovered: Expected '-' to be '+'.

  Message:
    Incorrect polarity discovered: Expected '-' to be '+'.

  Message:
    Incorrect polarity discovered: Expected '-' to be '+'.

141) llama3.3:70b-instruct-q5_K_M | causal translation testing | multiple feedback loops | extract 2 feedback loops with [+, +] polarities
  Message:
    Incorrect polarity discovered: Expected '-' to be '+'.

  Message:
    Incorrect polarity discovered: Expected '-' to be '+'.

Message:
    Incorrect polarity discovered: Expected '-' to be '+'.

142) llama3.3:70b-instruct-q5_K_M | causal translation testing | multiple feedback loops | extract 3 feedback loops with [-, -, +] polarities
  Message:
    Fake relationships found
    balacks --> (+) maxabizers, exemintes --> (-) frimbulators, marticatenes --> (+) frimbulators, marticatenes --> (-) funkados, refluppers --> (+) frimbulators, whatajigs --> (-) maxabizers, whoziewhats --> (-) maxabizers
    Ground Truth
    balacks --> (+) whoziewhats, exemintes --> (+) maxabizers, frimbulators --> (-) whatajigs, funkados --> (+) frimbulators, funkados --> (-) maxabizers, marticatenes --> (+) refluppers, maxabizers --> (+) funkados, maxabizers --> (+) marticatenes, refluppers --> (+) exemintes, whatajigs --> (+) balacks, whoziewhats --> (+) funkados: Expected 7 to be 0.

  Message:
    Incorrect polarity discovered: Expected '-' to be '+'.

  Message:
    Incorrect polarity discovered: Expected '-' to be '+'.

  Message:
    Incorrect polarity discovered: Expected '-' to be '+'.

  Message:
    Incorrect polarity discovered: Expected '-' to be '+'.

143) llama3.3:70b-instruct-q5_K_M | causal translation testing | single feedback loop | extract a reinforcing feedback loop with 8 variables
  Message:
    Fake relationships found
    Exemintes --> (-) Osc, Osc --> (+) Proptimatires
    Ground Truth
    auspongs --> (+) dominitoxings, dominitoxings --> (+) exemintes, exemintes --> (+) ocs, houtals --> (+) povals, ocs --> (+) proptimatires, povals --> (+) auspongs, priaries --> (+) houtals, proptimatires --> (+) priaries: Expected 2 to be 0.

  Message:
    Real relationships not found
    exemintes --> (+) ocs, ocs --> (+) proptimatires
    Ground Truth
    auspongs --> (+) dominitoxings, dominitoxings --> (+) exemintes, exemintes --> (+) ocs, houtals --> (+) povals, ocs --> (+) proptimatires, povals --> (+) auspongs, priaries --> (+) houtals, proptimatires --> (+) priaries: Expected 2 to be 0.

144) llama3.3:70b-instruct-q5_K_M | causal translation testing | multiple feedback loops | extract 5 feedback loops with [-, +, +, +, -] polarities
  Message:
    Fake relationships found
    illigents --> (-) frimbulators
    Ground Truth
    auspongs --> (+) dominitoxings, balacks --> (+) frimbulators, balacks --> (+) whoziewhats, dominitoxings --> (+) outrances, exemintes --> (+) ocs, frimbulators --> (-) whatajigs, funkados --> (+) maxabizers, houtals --> (+) priaries, houtals --> (-) povals, illigents --> (+) houtals, marticatenes --> (+) balacks, marticatenes --> (+) refluppers, maxabizers --> (+) marticatenes, ocs --> (+) proptimatires, outrances --> (+) illigents, povals --> (+) auspongs, priaries --> (+) houtals, priaries --> (+) marticatenes, proptimatires --> (+) priaries, refluppers --> (+) exemintes, whatajigs --> (+) balacks, whoziewhats --> (+) funkados: Expected 1 to be 0.

  Message:
    Real relationships not found
    houtals --> (+) priaries, illigents --> (+) houtals, priaries --> (+) marticatenes
    Ground Truth
    auspongs --> (+) dominitoxings, balacks --> (+) frimbulators, balacks --> (+) whoziewhats, dominitoxings --> (+) outrances, exemintes --> (+) ocs, frimbulators --> (-) whatajigs, funkados --> (+) maxabizers, houtals --> (+) priaries, houtals --> (-) povals, illigents --> (+) houtals, marticatenes --> (+) balacks, marticatenes --> (+) refluppers, maxabizers --> (+) marticatenes, ocs --> (+) proptimatires, outrances --> (+) illigents, povals --> (+) auspongs, priaries --> (+) houtals, priaries --> (+) marticatenes, proptimatires --> (+) priaries, refluppers --> (+) exemintes, whatajigs --> (+) balacks, whoziewhats --> (+) funkados: Expected 3 to be 0.

  Message:
    Incorrect polarity discovered: Expected '-' to be '+'.

  Message:
    Incorrect polarity discovered: Expected '-' to be '+'.

  Message:
    Incorrect polarity discovered: Expected '-' to be '+'.

  Message:
    Incorrect polarity discovered: Expected '-' to be '+'.

145) llama3.3:70b-instruct-q5_K_M | causal translation testing | single feedback loop | extract a reinforcing feedback loop with 3 variables
  Message:
    Incorrect polarity discovered: Expected '-' to be '+'.

  Message:
    Incorrect polarity discovered: Expected '-' to be '+'.

146) llama3.3:70b-instruct-q5_K_M | causal translation testing | single feedback loop | extract a balancing feedback loop with 3 variables
  Message:
    Incorrect polarity discovered: Expected '-' to be '+'.

147) llama3.3:70b-instruct-q5_K_M | causal translation testing | single feedback loop | extract a reinforcing feedback loop with 4 variables
  Message:
    Fake relationships found
    Marticatenes --> (-) Ref luppers
    Ground Truth
    funkados --> (+) maxabizers, marticatenes --> (+) refluppers, maxabizers --> (+) marticatenes, refluppers --> (+) funkados: Expected 1 to be 0.

  Message:
    Real relationships not found
    marticatenes --> (+) refluppers
    Ground Truth
    funkados --> (+) maxabizers, marticatenes --> (+) refluppers, maxabizers --> (+) marticatenes, refluppers --> (+) funkados: Expected 1 to be 0.

  Message:
    Incorrect polarity discovered: Expected '-' to be '+'.

148) llama3.3:70b-instruct-q5_K_M | causal translation testing | single feedback loop | extract a balancing feedback loop with 7 variables
  Message:
    Incorrect polarity discovered: Expected '-' to be '+'.

149) llama3.3:70b-instruct-q5_K_M | causal translation testing | single feedback loop | extract a reinforcing feedback loop with 5 variables
  Message:
    Fake relationships found
    Exemintes --> (-) Maxabizers, Marticatenes --> (+) Ocs, Marticatenes --> (-) Exemintes, Maxabizers --> (+) Exemintes, Maxabizers --> (-) Ocs, Maxabizers --> (-) Refluppers, Ocs --> (-) Exemintes, Refluppers --> (+) Maxabizers, Refluppers --> (+) Ocs
    Ground Truth
    exemintes --> (+) ocs, marticatenes --> (+) refluppers, maxabizers --> (+) marticatenes, ocs --> (+) maxabizers, refluppers --> (+) exemintes: Expected 9 to be 0.

  Message:
    Incorrect polarity discovered: Expected '-' to be '+'.

  Message:
    Incorrect polarity discovered: Expected '-' to be '+'.

Message:
    Incorrect polarity discovered: Expected '-' to be '+'.

150) llama3.3:70b-instruct-q5_K_M | causal translation testing | single feedback loop | extract a balancing feedback loop with 6 variables
    Message:
    Fake relationships found
    exemintes --> (-) proptimatires, ocs --> (+) priaries, priaries --> (-) refluppers, proptimatires --> (-) marticatenes, refluppers --> (+) ocs
    Ground Truth
    exemintes --> (+) ocs, marticatenes --> (-) refluppers, ocs --> (+) proptimatires, priaries --> (+) marticatenes, proptimatires --> (+) priaries, refluppers --> (+) exemintes: Expected 5 to be 0.

    Message:
    Incorrect polarity discovered: Expected '-' to be '+'.

    Message:
    Incorrect polarity discovered: Expected '-' to be '+'.

151) llama3.3:70b-instruct-q5_K_M | causal translation testing | multiple feedback loops | extract 5 feedback loops with [-, +, +, -, -] polarities
    Message:
    Fake relationships found
    flumplenooks --> (+) whatajigs, nimbuls --> (+) flumplenooks, ocs --> (+) proptants, proptants --> (+) nimbuls
    Ground Truth
    auspongs --> (+) dominitoxings, balacks --> (+) frimbulators, balacks --> (+) whoziewhats, dominitoxings --> (+) outrances, exemintes --> (+) ocs, frimbulators --> (-) whatajigs, funkados --> (+) maxabizers, houtals --> (+) priaries, houtals --> (-) povals, illigents --> (+) houtals, marticatenes --> (+) balacks, marticatenes --> (+) refluppers, maxabizers --> (+) marticatenes, ocs --> (+) proptimatires, outrances --> (+) illigents, povals --> (+) auspongs, priaries --> (+) marticatenes, priaries --> (-) houtals, proptimatires --> (+) priaries, refluppers --> (+) exemintes, whatajigs --> (+) balacks, whoziewhats --> (+) funkados: Expected 4 to be 0.

    Message:
    Real relationships not found
    auspongs --> (+) dominitoxings, dominitoxings --> (+) outrances, houtals --> (+) priaries, houtals --> (-) povals, illigents --> (+) houtals, ocs --> (+) proptimatires, outrances --> (+) illigents, povals --> (+) auspongs, priaries --> (+) marticatenes, priaries --> (-) houtals, proptimatires --> (+) priaries
    Ground Truth
    auspongs --> (+) dominitoxings, balacks --> (+) frimbulators, balacks --> (+) whoziewhats, dominitoxings --> (+) outrances, exemintes --> (+) ocs, frimbulators --> (-) whatajigs, funkados --> (+) maxabizers, houtals --> (+) priaries, houtals --> (-) povals, illigents --> (+) houtals, marticatenes --> (+) balacks, marticatenes --> (+) refluppers, maxabizers --> (+) marticatenes, ocs --> (+) proptimatires, outrances --> (+) illigents, povals --> (+) auspongs, priaries --> (+) marticatenes, priaries --> (-) houtals, proptimatires --> (+) priaries, refluppers --> (+) exemintes, whatajigs --> (+) balacks, whoziewhats --> (+) funkados: Expected 11 to be 0.

Message:
    Incorrect polarity discovered: Expected '-' to be '+'.

  Message:
    Incorrect polarity discovered: Expected '-' to be '+'.

  Message:
    Incorrect polarity discovered: Expected '-' to be '+'.

  Message:
    Incorrect polarity discovered: Expected '-' to be '+'.

  Message:
    Incorrect polarity discovered: Expected '-' to be '+'.

152) llama3.3:70b-instruct-q5_K_M | causal translation testing | multiple feedback loops | extract 3 feedback loops with [+, +, -] polarities
  Message:
    Incorrect polarity discovered: Expected '-' to be '+'.

  Message:
    Incorrect polarity discovered: Expected '-' to be '+'.

  Message:
    Incorrect polarity discovered: Expected '-' to be '+'.

  Message:
    Incorrect polarity discovered: Expected '-' to be '+'.

  Message:
    Incorrect polarity discovered: Expected '-' to be '+'.

153) llama3.3:70b-instruct-q5_K_M | causal translation testing | single feedback loop | extract a reinforcing feedback loop with 6 variables
  Message:
    Fake relationships found
    Exemintes --> (-) Priaries, Exemintes --> (-) Proptimatires, Marticatenes --> (+) Exemintes, Marticatenes --> (+) Ocs, Marticatenes --> (+) Priaries, Marticatenes --> (+) Proptimatires, Marticatenes --> (+) Ref luppers, Ocs --> (-) Priaries, Priaries --> (+) Ocs, Priaries --> (+) Proptimatires, Priaries --> (+) Ref luppers, Priaries --> (-) Exemintes, Ref luppers --> (+) Exemintes, Ref luppers --> (-) Ocs, Ref luppers --> (-) Priaries, Ref luppers --> (-) Proptimatires
    Ground Truth
    exemintes --> (+) ocs, marticatenes --> (+) refluppers, ocs --> (+) proptimatires, priaries --> (+) marticatenes, proptimatires --> (+) priaries, refluppers --> (+) exemintes: Expected 16 to be 0.

  Message:

Real relationships not found
marticatenes --> (+) refluppers, refluppers --> (+) exemintes
Ground Truth
exemintes --> (+) ocs, marticatenes --> (+) refluppers, ocs --> (+) proptimatires, priaries --> (+) marticatenes, proptimatires --> (+) priaries, refluppers --> (+) exemintes: Expected 2 to be 0.

Message:
Incorrect polarity discovered: Expected '-' to be '+'.

Message:
Incorrect polarity discovered: Expected '-' to be '+'.

154) llama3.3:70b-instruct-q5_K_M | conformance testing | can conform to the instruction include a maximum number of variables| for the case American Revolution
Message:
Too many variables: Variables are: Taxation, Anti-British Sentiment, Boston Massacre, Defiance Acts, Intolerable Acts, Colonial Identity: Expected 6 to be less than or equal 5.

155) llama3.3:70b-instruct-q5_K_M | conformance testing | can conform to the instruction include a maximum number of feedback loops| for the case American Revolution
Message:
Too many feedback loops: The number of feedback loops found was 5: Expected 5 to be less than or equal 4.

156) llama3.3:70b-instruct-q5_K_M | conformance testing | can conform to the instruction include a minimum number of feedback loops| for the case American Revolution
Message:
Too few feedback loops: The number of feedback loops found was 3: Expected 3 to be greater than or equal 8.

157) llama3.3:70b-instruct-q5_K_M | conformance testing | can conform to the instruction include a maximum number of feedback loops and a minimum number of variables| for the case American Revolution
Message:
Too many feedback loops: The number of feedback loops found was 6: Expected 6 to be less than or equal 4.

158) llama3.3:70b-instruct-q5_K_M | conformance testing | can conform to the instruction include a min number of feedback loops and a maximum number of variables| for the case American Revolution
Message:
Too few feedback loops: The number of feedback loops found was 4: Expected 4 to be greater than or equal 6.

159) deepseek-r1:32b | conformance testing | can conform to the instruction include a maximum number of feedback loops and a maximum number of variables| for the case American Revolution
   Message:
    Too many variables: Variables are: Taxation, Resentment, Boston Massacre, Hostilities Escalate, Boston Tea Party, Defiance against British Rule, French and Indian War Costs, British Demands for Repayment, Colonial Identity and Unity, American Revolution: Expected 10 to be less than or equal 5.

160) deepseek-r1:32b | conformance testing | can conform to the instruction include a minimum number of feedback loops and a minimum number of variables| for the case Road Rage
   Message:
    Too few feedback loops: The number of feedback loops found was 1: Expected 1 to be greater than or equal 6.

161) deepseek-r1:32b | conformance testing | can conform to the instruction include a minimum number of feedback loops and a minimum number of variables| for the case American Revolution
   Message:
    Too few feedback loops: The number of feedback loops found was 0: Expected 0 to be greater than or equal 6.

162) deepseek-r1:32b | conformance testing | can conform to the instruction include a min number of feedback loops and a maximum number of variables| for the case Road Rage
   Message:
    Too few feedback loops: The number of feedback loops found was 0: Expected 0 to be greater than or equal 6.

163) deepseek-r1:32b | conformance testing | can conform to the instruction include a maximum number of feedback loops and a maximum number of variables| for the case Road Rage
   Message:
    Too many variables: Variables are: Advanced technology, Reduction in human error, Urban design improvements, Reduction in traffic congestion, Public awareness campaigns, Improved anger management among drivers: Expected 6 to be less than or equal 5.

164) deepseek-r1:32b | conformance testing | can conform to the instruction include a minimum number of feedback loops| for the case American Revolution
   Message:
    Too few feedback loops: The number of feedback loops found was 0: Expected 0 to be greater than or equal 8.

165) deepseek-r1:32b | conformance testing | can conform to the instruction include a maximum number of variables| for the case American Revolution
   Message:

    Too many variables: Variables are: Taxation, Resistance, Boston Massacre, Boston Tea Party, Harsh Laws, French and Indian War Debt, Colonial Identity: Expected 7 to be less than or equal 5.

166) deepseek-r1:32b | conformance testing | can conform to the instruction include a maximum number of feedback loops and a minimum number of variables| for the case Road Rage
  Message:
    Too few variables: Variables are: Traffic Congestion, Road Rage, Psychological Stress, Learned Behavior: Expected 4 to be greater than or equal 5.

167) deepseek-r1:32b | conformance testing | can conform to the instruction include a minimum number of variables| for the case American Revolution
  Message:
    Too few variables: Variables are: French_Indian_War, Britishntaxes, Taxation, Boston_Massacre, Identity_Development, Boston_Tea_Party, Intolerable_Acts, Revolution_Outcome: Expected 8 to be greater than or equal 10.

168) deepseek-r1:32b | conformance testing | can conform to the instruction include a min number of feedback loops and a maximum number of variables| for the case American Revolution
  Message:
    Too few feedback loops: The number of feedback loops found was 0: Expected 0 to be greater than or equal 6.

169) deepseek-r1:32b | causal translation testing | single feedback loop | extract a balancing feedback loop with 2 variables
  Message:
    Fake relationships found
    balacks (B) --> (+) whoziewhats (W), whoziewhats (W) --> (+) balacks (B)
    Ground Truth
    balacks --> (-) whoziewhats, whoziewhats --> (+) balacks: Expected 2 to be 0.

  Message:
    Real relationships not found
    balacks --> (-) whoziewhats, whoziewhats --> (+) balacks
    Ground Truth
    balacks --> (-) whoziewhats, whoziewhats --> (+) balacks: Expected 2 to be 0.

170) deepseek-r1:32b | causal translation testing | single feedback loop | extract a reinforcing feedback loop with 8 variables
  Message:
    Incorrect polarity discovered: Expected '-' to be '+'.

  Message:
    Incorrect polarity discovered: Expected '-' to be '+'.

Message:
Incorrect polarity discovered: Expected '-' to be '+'.

Message:
Incorrect polarity discovered: Expected '-' to be '+'.

171) deepseek-r1:32b | causal translation testing | single feedback loop | extract a reinforcing feedback loop with 3 variables
Message:
Incorrect polarity discovered: Expected '-' to be '+'.

Message:
Incorrect polarity discovered: Expected '-' to be '+'.

172) deepseek-r1:32b | causal translation testing | single feedback loop | extract a balancing feedback loop with 4 variables
Message:
Fake relationships found
reflupppers --> (+) funkados
Ground Truth
funkados --> (-) maxabizers, marticatenes --> (+) refluppers, maxabizers --> (+) marticatenes, refluppers --> (+) funkados: Expected 1 to be 0.

Message:
Real relationships not found
refluppers --> (+) funkados
Ground Truth
funkados --> (-) maxabizers, marticatenes --> (+) refluppers, maxabizers --> (+) marticatenes, refluppers --> (+) funkados: Expected 1 to be 0.

Message:
Incorrect polarity discovered: Expected '-' to be '+'.

173) deepseek-r1:32b | causal translation testing | single feedback loop | extract a reinforcing feedback loop with 6 variables
Message:
Incorrect polarity discovered: Expected '-' to be '+'.

Message:
Incorrect polarity discovered: Expected '-' to be '+'.

174) deepseek-r1:32b | causal translation testing | single feedback loop | extract a balancing feedback loop with 3 variables

Message:
    Incorrect polarity discovered: Expected '-' to be '+'.

  Message:
    Incorrect polarity discovered: Expected '+' to be '-'.

175) deepseek-r1:32b | causal translation testing | single feedback loop | extract a balancing feedback loop with 6 variables
  Message:
    Fake relationships found
    proptomatires --> (-) priaries
    Ground Truth
    exemintes --> (+) ocs, marticatenes --> (-) refluppers, ocs --> (+) proptimatires, priaries --> (+) marticatenes, proptimatires --> (+) priaries, refluppers --> (+) exemintes: Expected 1 to be 0.

  Message:
    Real relationships not found
    proptimatires --> (+) priaries
    Ground Truth
    exemintes --> (+) ocs, marticatenes --> (-) refluppers, ocs --> (+) proptimatires, priaries --> (+) marticatenes, proptimatires --> (+) priaries, refluppers --> (+) exemintes: Expected 1 to be 0.

  Message:
    Incorrect polarity discovered: Expected '-' to be '+'.

176) deepseek-r1:32b | causal translation testing | single feedback loop | extract a reinforcing feedback loop with 4 variables
  Message:
    Incorrect polarity discovered: Expected '-' to be '+'.

177) deepseek-r1:32b | causal translation testing | single feedback loop | extract a reinforcing feedback loop with 7 variables
  Message:
    Incorrect polarity discovered: Expected '-' to be '+'.

  Message:
    Incorrect polarity discovered: Expected '-' to be '+'.

  Message:
    Incorrect polarity discovered: Expected '-' to be '+'.

  Message:
    Incorrect polarity discovered: Expected '-' to be '+'.

178) deepseek-r1:32b | causal translation testing | single feedback loop | extract a balancing feedback loop with 8 variables
  Message:
    Incorrect polarity discovered: Expected '-' to be '+'.

  Message:
    Incorrect polarity discovered: Expected '-' to be '+'.

  Message:
    Incorrect polarity discovered: Expected '-' to be '+'.

179) deepseek-r1:32b | causal translation testing | single feedback loop | extract a reinforcing feedback loop with 5 variables
  Message:
    Fake relationships found
    exemints --> (+) ocs, refluppers --> (-) exemints
    Ground Truth
    exemintes --> (+) ocs, marticatenes --> (+) refluppers, maxabizers --> (+) marticatenes, ocs --> (+) maxabizers, refluppers --> (+) exemintes: Expected 2 to be 0.

  Message:
    Real relationships not found
    exemintes --> (+) ocs, refluppers --> (+) exemintes
    Ground Truth
    exemintes --> (+) ocs, marticatenes --> (+) refluppers, maxabizers --> (+) marticatenes, ocs --> (+) maxabizers, refluppers --> (+) exemintes: Expected 2 to be 0.

  Message:
    Incorrect polarity discovered: Expected '-' to be '+'.

  Message:
    Incorrect polarity discovered: Expected '-' to be '+'.

180) deepseek-r1:32b | causal translation testing | multiple feedback loops | extract 5 feedback loops with [-, +, +, -, -] polarities
  Message:
    Fake relationships found
    woziewhats --> (+) funkados
    Ground Truth
    auspongs --> (+) dominitoxings, balacks --> (+) frimbulators, balacks --> (+) whoziewhats, dominitoxings --> (+) outrances, exemintes --> (+) ocs, frimbulators --> (-) whatajigs, funkados --> (+) maxabizers, houtals --> (+) priaries, houtals --> (-) povals, illigents --> (+) houtals, marticatenes --> (+) balacks, marticatenes --> (+) refluppers, maxabizers --> (+) marticatenes, ocs --> (+) proptimatires, outrances --> (+) illigents, povals --> (+) auspongs, priaries --> (+) marticatenes, priaries --> (-) houtals, proptimatires --> (+) priaries, refluppers --> (+) exemintes, whatajigs --> (+) balacks, whoziewhats --> (+) funkados: Expected 1 to be 0.

Message:
  Real relationships not found
  whoziewhats --> (+) funkados
  Ground Truth
  auspongs --> (+) dominitoxings, balacks --> (+) frimbulators, balacks --> (+) whoziewhats, dominitoxings --> (+) outrances, exemintes --> (+) ocs, frimbulators --> (-) whatajigs, funkados --> (+) maxabizers, houtals --> (+) priaries, houtals --> (-) povals, illigents --> (+) houtals, marticatenes --> (+) balacks, marticatenes --> (+) refluppers, maxabizers --> (+) marticatenes, ocs --> (+) proptimatires, outrances --> (+) illigents, povals --> (+) auspongs, priaries --> (+) marticatenes, priaries --> (-) houtals, proptimatires --> (+) priaries, refluppers --> (+) exemintes, whatajigs --> (+) balacks, whoziewhats --> (+) funkados: Expected 1 to be 0.

Message:
  Incorrect polarity discovered: Expected '-' to be '+'.

Message:
  Incorrect polarity discovered: Expected '-' to be '+'.

Message:
  Incorrect polarity discovered: Expected '-' to be '+'.

Message:
  Incorrect polarity discovered: Expected '-' to be '+'.

Message:
  Incorrect polarity discovered: Expected '+' to be '-'.

Message:
  Incorrect polarity discovered: Expected '-' to be '+'.

Message:
  Incorrect polarity discovered: Expected '+' to be '-'.

Message:
  Incorrect polarity discovered: Expected '-' to be '+'.

Message:
  Incorrect polarity discovered: Expected '-' to be '+'.

Message:
  Incorrect polarity discovered: Expected '-' to be '+'.

Message:
  Incorrect polarity discovered: Expected '-' to be '+'.

181) deepseek-r1:32b | causal translation testing | single feedback loop | extract a balancing feedback loop with 7 variables
  Message:

Incorrect polarity discovered: Expected '-' to be '+'.

  Message:
    Incorrect polarity discovered: Expected '-' to be '+'.

  Message:
    Incorrect polarity discovered: Expected '-' to be '+'.

  Message:
    Incorrect polarity discovered: Expected '+' to be '-'.

182) deepseek-r1:32b | causal translation testing | multiple feedback loops | extract 5 feedback loops with [-, +, +, +, -] polarities
  Message:
    Fake relationships found
    Povals --> (+) Balacks
    Ground Truth
    auspongs --> (+) dominitoxings, balacks --> (+) frimbulators, balacks --> (+) whoziewhats, dominitoxings --> (+) outrances, exemintes --> (+) ocs, frimbulators --> (-) whatajigs, funkados --> (+) maxabizers, houtals --> (+) priaries, houtals --> (-) povals, illigents --> (+) houtals, marticatenes --> (+) balacks, marticatenes --> (+) refluppers, maxabizers --> (+) marticatenes, ocs --> (+) proptimatires, outrances --> (+) illigents, povals --> (+) auspongs, priaries --> (+) houtals, priaries --> (+) marticatenes, proptimatires --> (+) priaries, refluppers --> (+) exemintes, whatajigs --> (+) balacks, whoziewhats --> (+) funkados: Expected 1 to be 0.

  Message:
    Real relationships not found
    auspongs --> (+) dominitoxings, dominitoxings --> (+) outrances, illigents --> (+) houtals, outrances --> (+) illigents, povals --> (+) auspongs
    Ground Truth
    auspongs --> (+) dominitoxings, balacks --> (+) frimbulators, balacks --> (+) whoziewhats, dominitoxings --> (+) outrances, exemintes --> (+) ocs, frimbulators --> (-) whatajigs, funkados --> (+) maxabizers, houtals --> (+) priaries, houtals --> (-) povals, illigents --> (+) houtals, marticatenes --> (+) balacks, marticatenes --> (+) refluppers, maxabizers --> (+) marticatenes, ocs --> (+) proptimatires, outrances --> (+) illigents, povals --> (+) auspongs, priaries --> (+) houtals, priaries --> (+) marticatenes, proptimatires --> (+) priaries, refluppers --> (+) exemintes, whatajigs --> (+) balacks, whoziewhats --> (+) funkados: Expected 5 to be 0.

  Message:
    Incorrect polarity discovered: Expected '-' to be '+'.

  Message:
    Incorrect polarity discovered: Expected '-' to be '+'.

  Message:
    Incorrect polarity discovered: Expected '-' to be '+'.

  Message:
    Incorrect polarity discovered: Expected '+' to be '-'.

Message:
    Incorrect polarity discovered: Expected '-' to be '+'.

  Message:
    Incorrect polarity discovered: Expected '+' to be '-'.

  Message:
    Incorrect polarity discovered: Expected '-' to be '+'.

  Message:
    Incorrect polarity discovered: Expected '-' to be '+'.

  Message:
    Incorrect polarity discovered: Expected '-' to be '+'.

183) deepseek-r1:32b | causal translation testing | multiple feedback loops | extract 3 feedback loops with [+, +, -] polarities
  Message:
    Fake relationships found
    Marticatenes --> (+) Ref luppers, Ref luppers --> (-) Exemintes
    Ground Truth
    balacks --> (+) whoziewhats, exemintes --> (+) maxabizers, frimbulators --> (+) whatajigs, funkados --> (+) frimbulators, funkados --> (+) maxabizers, marticatenes --> (+) refluppers, maxabizers --> (+) funkados, maxabizers --> (-) marticatenes, refluppers --> (+) exemintes, whatajigs --> (+) balacks, whoziewhats --> (+) funkados: Expected 2 to be 0.

  Message:
    Real relationships not found
    marticatenes --> (+) refluppers, refluppers --> (+) exemintes
    Ground Truth
    balacks --> (+) whoziewhats, exemintes --> (+) maxabizers, frimbulators --> (+) whatajigs, funkados --> (+) frimbulators, funkados --> (+) maxabizers, marticatenes --> (+) refluppers, maxabizers --> (+) funkados, maxabizers --> (-) marticatenes, refluppers --> (+) exemintes, whatajigs --> (+) balacks, whoziewhats --> (+) funkados: Expected 2 to be 0.

  Message:
    Incorrect polarity discovered: Expected '-' to be '+'.

  Message:
    Incorrect polarity discovered: Expected '-' to be '+'.

  Message:
    Incorrect polarity discovered: Expected '-' to be '+'.

  Message:
    Incorrect polarity discovered: Expected '-' to be '+'.

184) deepseek-r1:32b | causal translation testing | multiple feedback loops | extract 2 feedback loops with [-, +] polarities
  Message:
    Real relationships not found
    balacks --> (+) frimbulators, balacks --> (+) whoziewhats, funkados --> (+) maxabizers, marticatenes --> (+) refluppers, maxabizers --> (+) marticatenes, refluppers --> (+) balacks, whatajigs --> (+) balacks, whoziewhats --> (+) funkados
  Ground Truth
    balacks --> (+) frimbulators, balacks --> (+) whoziewhats, frimbulators --> (-) whatajigs, funkados --> (+) maxabizers, marticatenes --> (+) refluppers, maxabizers --> (+) marticatenes, refluppers --> (+) balacks, whatajigs --> (+) balacks, whoziewhats --> (+) funkados: Expected 8 to be 0.

185) deepseek-r1:32b | causal translation testing | multiple feedback loops | extract 3 feedback loops with [-, -, +] polarities
  Message:
    Incorrect polarity discovered: Expected '-' to be '+'.

  Message:
    Incorrect polarity discovered: Expected '-' to be '+'.

  Message:
    Incorrect polarity discovered: Expected '-' to be '+'.

  Message:
    Incorrect polarity discovered: Expected '-' to be '+'.

186) deepseek-r1:32b | causal translation testing | multiple feedback loops | extract 2 feedback loops with [+, +] polarities
  Message:
    Incorrect polarity discovered: Expected '-' to be '+'.

  Message:
    Incorrect polarity discovered: Expected '-' to be '+'.

  Message:
    Incorrect polarity discovered: Expected '-' to be '+'.

  Message:
    Incorrect polarity discovered: Expected '-' to be '+'.

  Message:
    Incorrect polarity discovered: Expected '-' to be '+'.

187) deepseek-r1:32b | causal translation testing | single relationship | extract a reinforcing relationship down
  Message:
    Real relationships not found
    frimbulators --> (+) whatajigs
    Ground Truth
    frimbulators --> (+) whatajigs: Expected 1 to be 0.

188) deepseek-r1:32b | causal translation testing | single relationship | extract a reinforcing relationship up
  Message:
    Fake relationships found
    Whatajigs --> (+) Frimbulators
    Ground Truth
    frimbulators --> (+) whatajigs: Expected 1 to be 0.